\documentclass[10pt,journal,compsoc]{IEEEtran}

\ifCLASSOPTIONcompsoc
  \usepackage[nocompress]{cite}
\else
  \usepackage{cite}
\fi

\ifCLASSINFOpdf
\else
\fi

\usepackage{times}
\usepackage{epsfig}
\usepackage{graphicx}
\usepackage{amsmath}
\usepackage{amssymb}
\usepackage{comment}
\usepackage{tikz}
\usepackage{url}

\definecolor{gold}{HTML}{BD820B}
\definecolor{silver}{HTML}{909090}
\definecolor{bronze}{HTML}{9A5F26}
\newcommand*\circledd[1]{\tikz[baseline=(char.base)]{
            \node[shape=circle,draw,inner sep=0.15pt] (char) {#1};}}           
\newcommand{\first}[1]{%
    {\raisebox{0.8pt}{\footnotesize \color{gold} \circledd{1}}\hspace{3.5pt}{\bf #1}}%
}
\newcommand{\second}[1]{%
    {\raisebox{0.8pt}{\footnotesize \color{silver} \circledd{2}}\hspace{3.5pt}#1}%
}
\newcommand{\third}[1]{%
    {\raisebox{0.8pt}{\footnotesize \color{bronze} \circledd{3}}\hspace{3.5pt}#1}%
}

\graphicspath{{./figures/}}

\begin{document}

\title{A Discriminative Single-Shot Segmentation Network for Visual Object Tracking}

\author{Alan~Lukežič$^1$, Jiří Matas$^2$, Matej~Kristan$^1$ \\
  {\small $^1$Faculty of Computer and Information Science, University of Ljubljana, Slovenia} \\
  {\small $^2$Faculty of Electrical Engineering, Czech Technical University in Prague, Czech Republic} \\
  {\tt\small alan.lukezic@fri.uni-lj.si}
}

\IEEEtitleabstractindextext{%
\begin{abstract}
Template-based discriminative trackers are currently the dominant tracking paradigm due to their robustness, but are restricted to bounding box tracking and a limited range of transformation models, which reduces their localization accuracy.
We propose a discriminative single-shot segmentation tracker -- D3S$_2$, which narrows the gap between visual object tracking and video object segmentation. 
A single-shot network applies two target models with complementary geometric properties, one invariant to a broad range of transformations, including non-rigid deformations, the other assuming a rigid object to simultaneously achieve robust online target segmentation. The overall tracking reliability is further increased by decoupling the object and feature scale estimation.
Without per-dataset finetuning, and trained only for segmentation as the primary output, D3S$_2$ outperforms all published trackers on the recent short-term tracking benchmark VOT2020 and performs very close to the state-of-the-art trackers on the GOT-10k, TrackingNet, OTB100 and LaSoT.
D3S$_2$ outperforms the leading segmentation tracker SiamMask on video  object segmentation benchmarks and performs on par with top video object segmentation algorithms.
\end{abstract}

}

\maketitle

\IEEEdisplaynontitleabstractindextext

%

\IEEEraisesectionheading{\section{Introduction}\label{sec:introduction}}

\IEEEPARstart{V}{isual} object tracking is one of core computer vision problems. 
The most common formulation considers the task of reporting target location in each frame of the video given a single training image. Currently, the dominant tracking paradigm, performing best in evaluations~\cite{kristan_vot2017,kristan_vot2018}, 
is correlation bounding box tracking~\cite{DanelljanCVPR2017, danelljan_eccv2018_updt, Lukezic_CVPR_2017,siamfc_eccv16,dasiamrpn_eccv2018,siamrpn_cvpr2018} where the target represented by a  multi-channel rectangular template 
is localized by cross-correlation between the template and a search region. 

\begin{figure}[!t]
\centering
\includegraphics[width=1\linewidth]{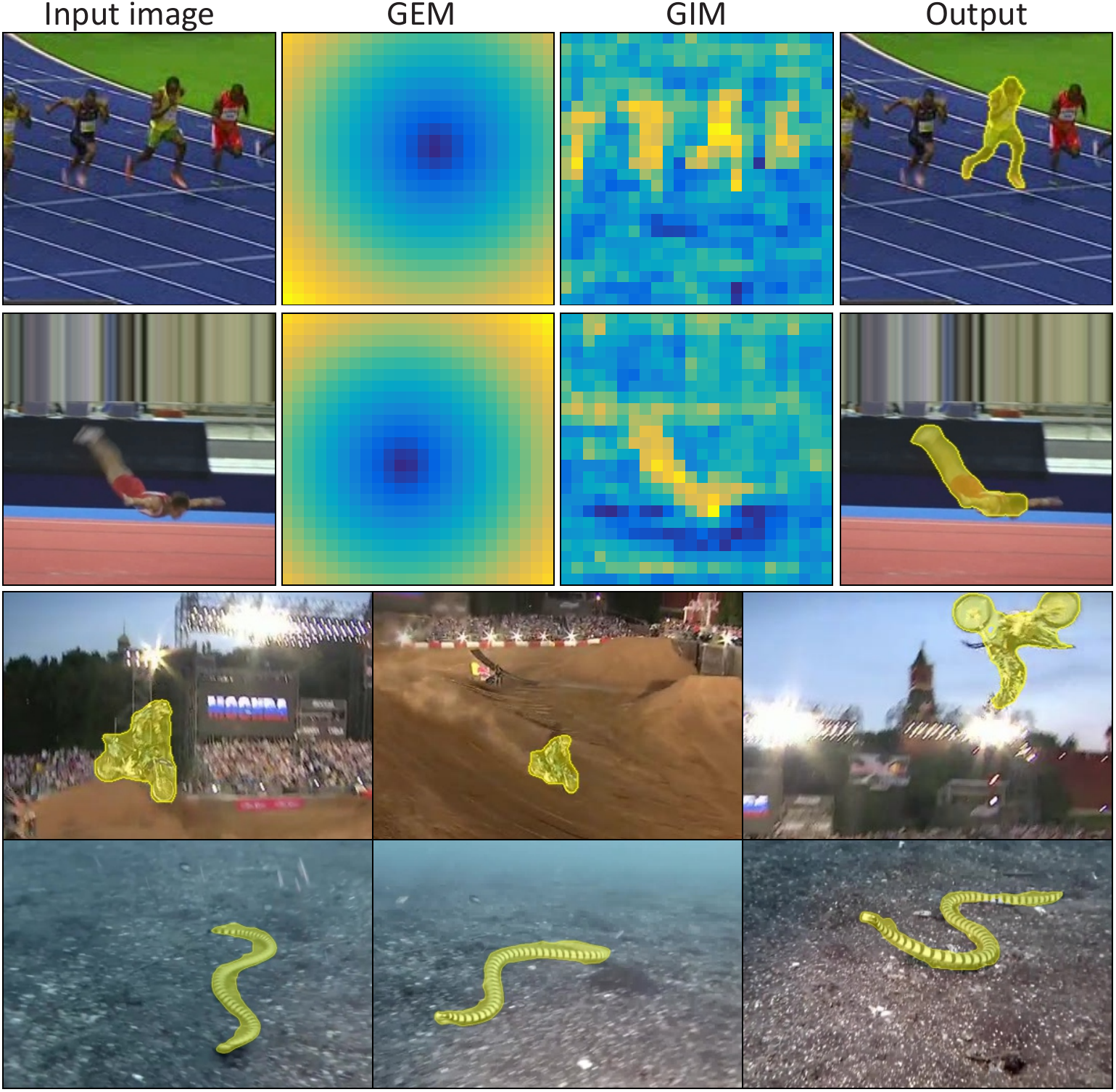}
\caption{
The D3S$_2$ tracker represents the target by two models with complementary geometric properties, one invariant to a broad range of transformations, including non-rigid deformations (GIM - geometrically invariant model), the other assuming a rigid object with motion well approximated by an euclidean transformation (GEM - geometrically constrained Euclidean model). 
The D3S$_2$, exploiting the complementary strengths of GIM and GEM, provides both state-of-the-art localisation and accurate segmentation, even in the presence of substantial deformation.
}
\label{fig:intro_figure}
\end{figure}  

State-of-the-art template-based trackers apply an efficient brute-force search for target localization. 
Such strategy is appropriate for low-dimensional transformations like translation and scale change, but becomes inefficient for more general situations e.g. such that induce an aspect ratio change and rotation. 
As a compromise, modern trackers combine approximate exhaustive search with sampling and/or bounding box refinement/regression networks~\cite{atom_cvpr19,siamrpn_cvpr2019} for aspect ratio estimation.
However, these approaches are restricted to axis-aligned rectangles. 

Estimation of high-dimensional template-based transformation is unreliable when a bounding box is a poor approximation of the target~\cite{lukezic_dpt}.
This is common --  consider e.g. elongated, rotating, deformable objects, or a person with spread out hands. 
In these cases, the most accurate and well-defined target location model is a binary per-pixel segmentation mask.
If such output is required, tracking becomes the video object segmentation task recently popularized by DAVIS~\cite{davis16,davis17} and YoutubeVOS~\cite{yt_vos2018} challenges. 

Unlike in tracking, video object segmentation challenges typically consider large targets observed for less than 100 frames with low background distractor presence. 
Top video object segmentation approaches thus fare poorly in short-term tracking scenarios~\cite{kristan_vot2018} where the target covers a fraction of the image, substantially changes its appearance over a longer period and moves over a cluttered background. 
Best trackers apply visual model adaptation, but in the case of segmentation errors it leads to an irrecoverable tracking failure~\cite{dat_cvpr15}. 
Because of this, in the past, segmentation has played only an auxiliary role in template-based trackers~\cite{staple_cvpr2016}, constrained DCF learning~\cite{Lukezic_CVPR_2017} and tracking by 3D model construction~\cite{otr_cvpr19}.

Recently, the SiamRPN~\cite{siamrpn_cvpr2018} tracker has been extended to address target localization and segmentation within a common framework called SiamMask~\cite{siammask_cvpr19}. 
The processing pipeline in SiamMask is split into two stages -- the first stage localizes the bounding box by a SiamRPN branch and then another branch estimates the segmentation mask by only considering the features within the bounding box. 
Splitting the processing pipeline into two separate stages misses the opportunity to treat localization and segmentation in an end-to-end framework. 
In addition, a fixed template is used that cannot be discriminatively adapted to the changing scene.

We propose a new single-shot discriminative segmentation tracker, D3S$_2$, that addresses the above-mentioned limitations. 
The target is encoded by two discriminative visual models -- one is adaptive and highly discriminative, but geometrically constrained to an Euclidean motion (GEM), while the other is invariant to broad range of transformation (GIM, geometrically invariant model), see Figure~\ref{fig:intro_figure}. 

GIM sacrifices spatial relations to allow target localization under significant deformation. On the other hand, GEM predicts only position, but discriminatively adapts to the target and acts as a selector between possibly multiple target segmentations inferred by GIM. 
In contrast to related trackers~\cite{siammask_cvpr19,siamrpn_cvpr2019,atom_cvpr19,dimp_iccv19}, the primary output of D3S$_2$ is a segmentation map computed in a single pass through the network, which is trained end-to-end for segmentation only (Figure~\ref{fig:architecture-overview}). 

Tracking reliability depends significantly on the target scale estimation accuracy, since the latter directly affects the spatial scale of the extracted features. A standard approach in state-of-the-art is to approximate the scale by the predicted bounding box, which commonly corresponds to the visible parts of the target. Such approximation is invalid in many cases. For example, during a partial occlusion, the visible part of the target becomes smaller, but the inherent target size and the feature scale, may not change.
Mismatching the scale leads to inaccurate model updates, unreliable target localization in the search region and eventual tracking failure. 
To address this issue, D3S$_2$ decouples the target inherent size (referred to the target scale in this paper) from the 
visible size (used for reporting target location in the image). The target scale is robustly predicted within the single-shot pipeline from the estimated segmentation mask, the target template and the appearance in the current frame, thus improving the feature extraction accuracy and tracking reliability.
The target location in the current image is reported as the predicted segmentation mask or by fitting a bounding box to the mask if required by the evaluation protocol.

A preliminary version of the tracker was published in~\cite{Lukezic_CVPR_2020} under the name D3S. 
The tracker presented in this paper extends the conference version by (i) the scale estimation module SEM, (ii) trainable similarity computation and (iii) channel attention in GIM and by (iv) using a stronger backbone and (v) a higher number of features in GEM. 
To avoid confusion with the preliminary version, we denote the new tracker as D3S$_2$. 
In addition to methodological improvements, we include the following new experiments: 
(i) a new experiment which exposes the ambiguity of manually annotated bounding boxes in datasets, 
(ii) segmentation-based annotations for ten OTB sequences for additional segmentation performance insights, 
(iii) an extended ablation study and 
(iv) additional qualitative analysis.

D3S$_2$ outperforms all published state-of-the-art trackers on the segmentation-based tracking dataset VOT2020~\cite{kristan_vot2020}, while ranking second among all participated trackers on the challenge. 
On the bounding box-based short-term tacking datasets, D3S$_2$ outperforms most state-of-the-art trackers (\cite{got10k,muller_trackingnet}) or performs on par~\cite{otb_pami2015} and performs close to the state-of-the-art on a long-term bounding-box tracking dataset~\cite{lasot_cvpr19} despite being a short-term tacker not trained for bounding box tracking. 
In video object segmentation benchmarks~\cite{davis16,davis17}, D3S$_2$ outperforms the leading segmentation tracker~\cite{siammask_cvpr19} and performs on par with top video object segmentation algorithms (often tuned to a specific domain), yet running orders of magnitude faster than most of them. 
Note that D3S$_2$ is not re-trained for different benchmarks -- 
a single pre-trained version shows remarkable generalization ability and versatility%
\footnote{A PyTorch implementation will be made publicly available on GITHUB.}.

\section{Related Work}  \label{sec:related_work}

Robust localization depends on the discrimination capability between the target and the background distractors. 
This property has been studied in depth in discriminative template trackers called discriminative correlation filters (DCF)~\cite{bolme2010visual}. The template learning is formulated as a (possibly nonlinear) ridge regression problem and solved by circular correlation~\cite{bolme2010visual,danelljan2014adaptive,henriques2015tracking,samf_eccv2014}.
Extensive effort has been invested by the research community to extend the DCF robustness by increasing the training and detection range via regularization~\cite{danelljan_eccv2016_ccot} and  masking~\cite{galoogahi_multi_channel_correlation} and by patch sampling~\cite{muller_context_aware_cvpr2017}.

While trackers based purely on color segmentation~\cite{comanichu_kernel_pami2003,dat_cvpr15} are inferior to DCFs, segmentation has been used for improved DCF tracking of non-rectangular targets~\cite{staple_cvpr2016,lukezic_dpt}. 
Lukežič et al.~\cite{Lukezic_CVPR_2017} used color segmentation to constrain DCF learning and proposed a real-time tracker with hand-crafted features which achieved performance comparable to trackers with deep features. 
The method was extended to long-term~\cite{Lukezic_ACCV_2018} and RGB-depth tracking~\cite{otr_cvpr19} using color and depth segmentation. 
Further improvements in DCF tracking considered deep features: Danelljan et al.~\cite{DanelljanCVPR2017} used features pre-trained for detection and Valmadre et al.~\cite{Valmadre_2017_CVPR} proposed pre-training features for DCF localization. 
Recently Danelljan et al.~\cite{atom_cvpr19} proposed a deep DCF training using backpropagation, 
while Bhat et al.~\cite{dimp_iccv19} extended it to the end-to-end trainable tracking architecture.

Another class of trackers, called Siamese trackers~\cite{siamfc_eccv16,tao2016sint,sa_siam_cvpr2018}, has evolved in direction of generative templates. 
Siamese trackers apply a backbone pre-trained offline with general targets such that object-background discrimination is maximized by correlation between the search region and target template extracted in the first frame~\cite{siamfc_eccv16}.
The template and the backbone are fixed during tracking, leading to an excellent real-time performance~\cite{kristan_vot2018}. 
Several multi-stage Siamese extensions have been proposed. These include addition of region proposal networks for improved target localization accuracy~\cite{siamrpn_cvpr2018,siamrpn_cvpr2019} and addition of segmentation branches~\cite{siammask_cvpr19} for accurate target segmentation. 
Voigtlaender et al.~\cite{siamrcnn_cvpr2020}, proposed a siamese two-stage tracking architecture, which combines a siamese tracker and a detector based on region proposal network. A common drawback of the siamese trackers is that template update by standard temporal averaging leads to reduced performance. This was recently addressed by proposing a template adaptation technique based on backpropagation~\cite{gradnet_iccv2019}.
Recent works~\cite{siamban_cvpr2020,siamcar_cvpr2020} have shown benefits of replacing the region proposal networks by modules inspired by anchor-free object detection networks~\cite{fcos_iccv2019}.

Segmentation of moving objects is a central problem in the emerging field of video object segmentation (VOS)~\cite{davis16,yt_vos2018}. 
Most recent works~\cite{feelvos_cvpr2019,osvos_cvpr2017,onavos_bmvc2017,favos_cvpr2018,osmn_cvpr2018} achieve impressive results, but involve large deep networks, which often require finetuning and are slow.
Hu et al.~\cite{videomatch_eccv2018} and Chen et al.~\cite{blazingly_fast_cvpr18} concurrently proposed segmentation by matching features extracted in the first frame, which considerably reduces the processing time. 
Further advancements of the VOS field focus mostly on target representation and online adaptation. 
Johnander et al.~\cite{Johnander_CVPR2019} proposed a network architecture that learns a generative model of target and background feature distributions and runs close to real-time. 
Recently, Robinson et al.~\cite{robinson_cvpr2020} adopted the optimizer from~\cite{atom_cvpr19} and used the estimated target module to compute the low-resolution segmentation. An upsampling module was proposed to predict a final high-resolution segmentation mask. 
Bhat et al.~\cite{bhat_eccv2020} proposed a differentiable VOS architecture for predicting a target parametric model by minimizing the segmentation error at model initialization, allows online adaptation and substantially improves the segmentation accuracy.
The VOS task, however, considers segmentation of large objects with limited appearance changes in short videos. Thus, these methods fare poorly on the visual object tracking task with small, fast moving objects. 
The work proposed in this paper aims at narrowing the gap between visual object tracking and video object segmentation.

\section{Discriminative segmentation network}  \label{sec:methods}

\begin{figure}[!t]
\begin{center}
	\includegraphics[width=\linewidth]{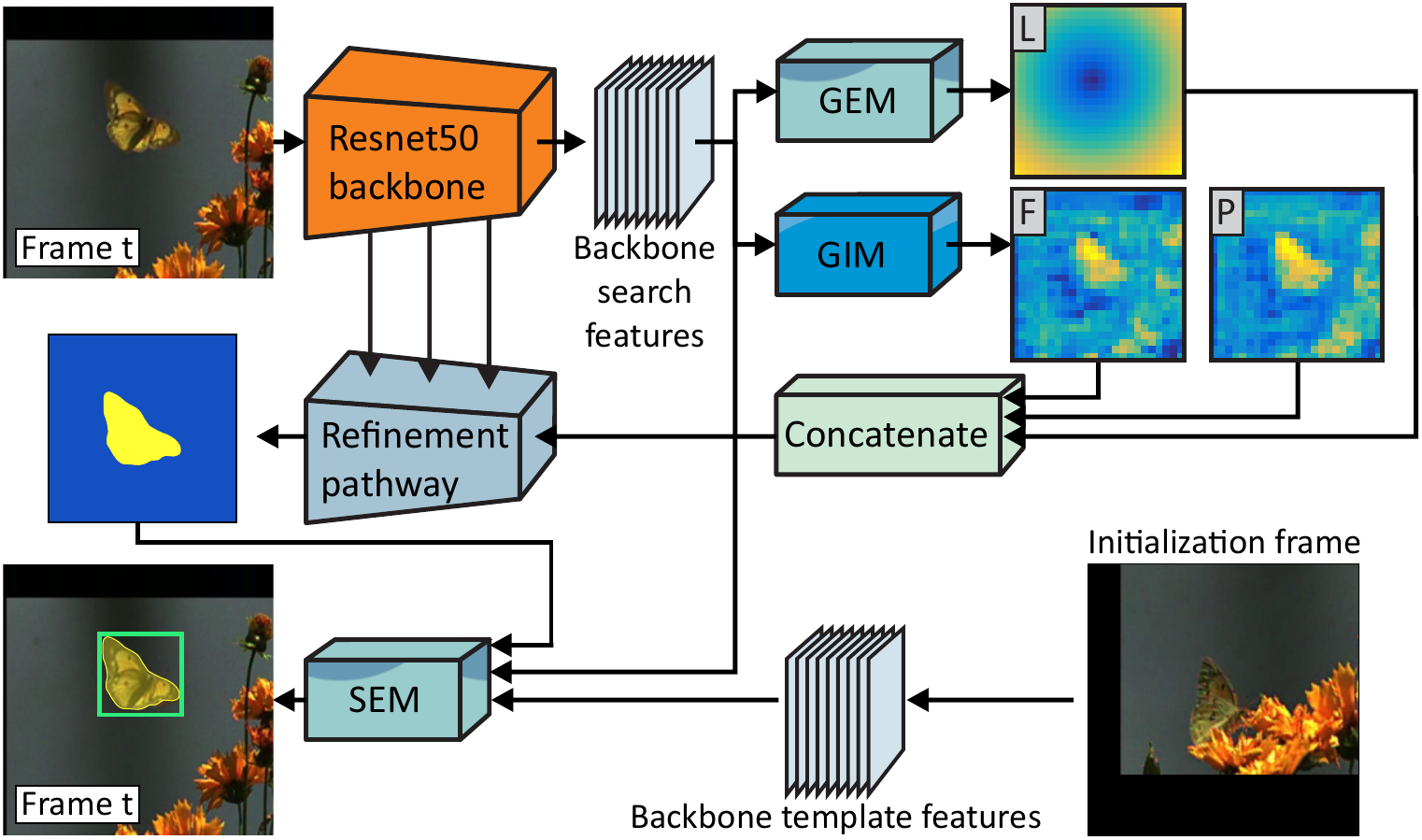}
\end{center}
   \caption{The D3S$_2$ segmentation architecture. The backbone features are  processed by the GEM and GIM pathways,
   producing the target location ($\mathbf{L}$), foreground similarity ($\mathbf{F}$) and target posterior ($\mathbf{P}$) channels.
   The three channels are concatenated and refined into a detailed segmentation mask, which is used in SEM for robust target size estimation.} 
\label{fig:architecture-overview} 
\end{figure}

The D3S$_2$ architecture is summarized in Figure~\ref{fig:architecture-overview}.
Two models are used to robustly cope with target appearance changes and background discrimination: a geometrically invariant model (GIM) presented in Section~\ref{sec:sum}, which is initialized from the target template, and a
geometrically constrained Euclidean model (GEM) presented in Section~\ref{sec:scm}, which adapts online to target appearance changes. 
These models process the input in parallel pathways and produce several coarse target presence channels, which are fused into a detailed segmentation mask by a refinement pathway described in Section~\ref{sec:refinement}. 
Finally, the target scale is estimated from the mask, model template and the search region by the scale estimation module (SEM) presented in Section~\ref{sec:scale-change-module}. 
The template is used at an early stage of the D3S architecture (in GIM) as well as at the late stage (SEM) -- D3S thus considers early as well as late template feature fusion for accurate target localization. 
The following subsections detail the architecture.

\subsection{Geometrically invariant model pathway}  \label{sec:sum}

Accurate segmentation of a deformable target requires loose spatial constraints in the discriminative model. 
Our geometrically invariant model (GIM) thus removes all positional information and is composed of two 
sets of deep feature vectors corresponding to the target and the background, i.e., $\mathbf{X}_\mathrm{GIM} = \{ \mathbf{X}^{F}, \mathbf{X}^{B}\}$. 

\begin{figure}[!h]
\begin{center}
	\includegraphics[width=0.6\linewidth]{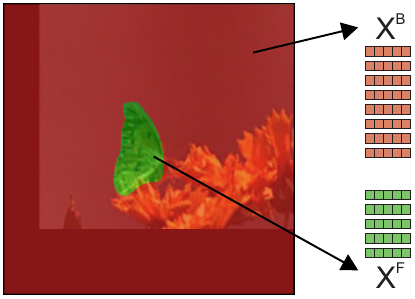}
\end{center}
   \caption{Target and background feature vectors in GIM are extracted in the first frame at pixel locations corresponding to the target ($\mathbf{X}^{F}$) and from the immediate neighborhood for the background ($\mathbf{X}^{B}$).}
\label{fig:gim-init-extraction} 
\end{figure}

The pre-trained backbone features are first processed by a $1\times 1$ convolutional layer to reduce their dimensionality to 64 channels, which is followed by a $3 \times 3$ convolutional layer (a ReLU is placed after each convolutional layer). 
Both these layers are adjusted in the network training stage to produce optimal features for segmentation. 
The target/background models are created in the first frame by extracting the segmentation feature vectors at pixel locations corresponding to the target ($\mathbf{X}^{F}$) and from the immediate neighborhood for the background ($\mathbf{X}^{B}$).
The extraction process is shown in Figure~\ref{fig:gim-init-extraction}.

\begin{figure}[!t]
\begin{center}
	\includegraphics[width=\linewidth]{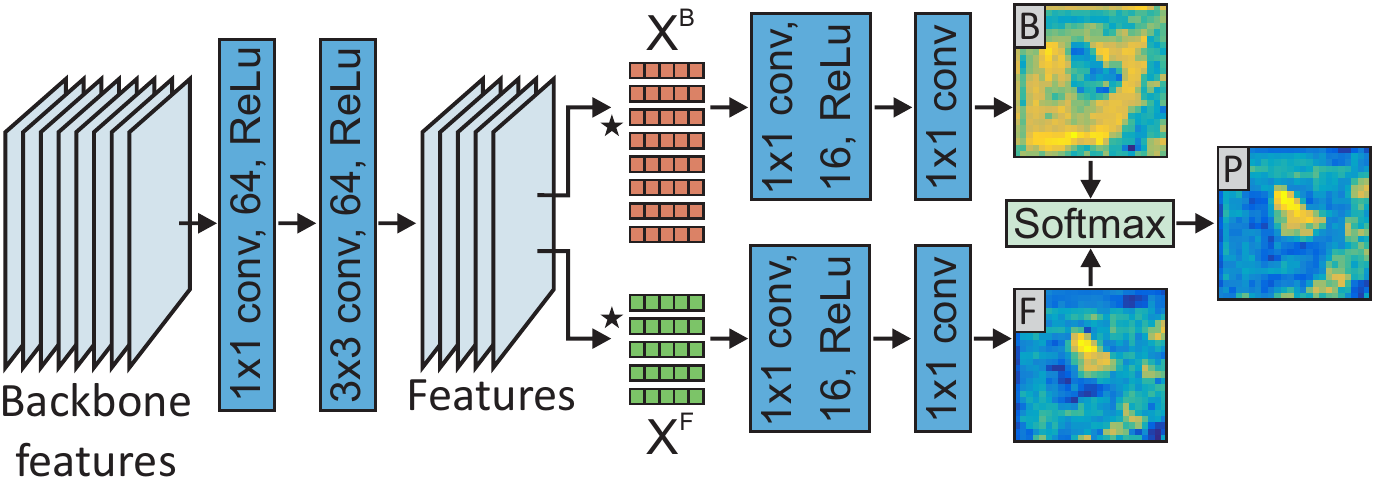}
\end{center}
   \caption{GIM -- the geometrically invariant model -- features are matched to the features in the foreground-background model $\{ \mathbf{X}^{F}, \mathbf{X}^{B}\}$ to obtain the target ($\mathbf{F}$) and background ($\mathbf{B}$) similarity channels.
   The posterior channel ($\mathbf{P}$) is the  softmax of $\mathbf{F}$ and $\mathbf{B}$.} 
\label{fig:sum-architecture} 
\end{figure}

During tracking, the pixel-level features extracted from the search region are compared to those of GIM ($\mathbf{X}_\mathrm{GIM}$) to compute foreground and background similarity channels $\mathbf{F}$ and $\mathbf{B}$ following~\cite{videomatch_eccv2018}. 
Specifically, for the $\mathbf{F}$ channel computation, each feature $\mathbf{y}_i$ extracted at pixel $i$ is compared to all features $\mathbf{x}_j^F \in \mathbf{X}^{F}$ by a normalized dot product
\begin{equation}  \label{eq:sum-similarity}
    s_{ij}^{F}(\mathbf{y}_i, \mathbf{x}_j^F) = \langle \tilde{\mathbf{y}}_i, \tilde{\mathbf{x}}_j^F \rangle,
\end{equation}
where $\tilde{(\cdot )}$ indicates an $L_2$ normalization. 
The similarity scores computed at pixel indexed by $i$ are sorted\footnote{Note that we use sorting which allows gradient backpropagation in the end-to-end network pre-training, even though the sorting operation is not differentiable on its own} in a descending order and passed to a multi-layer perceptron (MLP) for decoding into a single similarity score $\mathbf{F}_i$, i.e.,
\begin{equation}  \label{eq:fg-similarity}
    \mathbf{F}_i = \mathrm{MLP}( \mathrm{sort} \{s_{ij}^{F}\}_{j=1:N_F}).
\end{equation}
Note that various methods could be used to transform the set of per-pixel similarities $s^{F}_{ij}$ into a single foreground similarity score $\mathbf{F}_i$. 
We use the MLP as a good compromise between simplicity and flexibility to learn the optimal prototype set matching. In particular, we apply a two-layer MLP with $N_F$ input nodes, 16 hidden neurons and a single output.
Computation of the background similarity channel $\mathbf{B}$ follows the same principle, but with similarities computed with the background model feature vectors, i.e., $\mathbf{x}_j^B \in \mathbf{X}^{B}$. 
Finally, a softmax layer is applied to produce a target posterior channel $\mathbf{P}$.
The GIM pathway architecture is shown in Figure~\ref{fig:sum-architecture}.

\subsection{Geometrically constrained model pathway}  \label{sec:scm}

While GIM produces an excellent target-background separation, it cannot well distinguish the target from similar instances, leading to a reduced robustness (see Figure~\ref{fig:intro_figure}, first line). 
Robust localization, however, is a well-established quality of the discriminative correlation filters. 
Although they represent the target by a geometrically constrained model (i.e., a rectangular filter), efficient techniques developed to adapt to the target discriminative features~\cite{danelljan_eccv2016_ccot,Lukezic_CVPR_2017,atom_cvpr19} allow tracking reliably under considerable appearance changes.

We thus employ a recent deep DCF formulation~\cite{atom_cvpr19} in the geometrically constrained Euclidean model (GEM) pathway. 
The backbone features are first reduced to 256 channels by $1 \times 1$ convolutional layer. 
The reduced features are correlated by a 256 channel DCF followed by a PeLU nonlinearity~\cite{pelu_icmla17}. The reduction layer and DCF are trained by an efficient backprop formulation (see~\cite{atom_cvpr19} for details).  

\begin{figure}[!t]
\begin{center}
	\includegraphics[width=\linewidth]{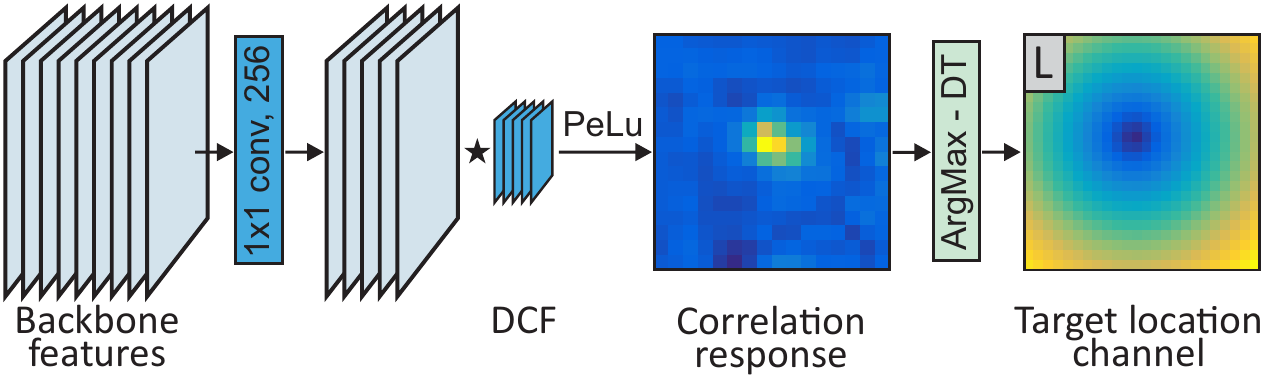}
\end{center}
   \caption{GEM -- the geometrically constrained Euclidean model -- reduces the backbone features dimensionality and correlates them with a DCF. 
   The target localisation channel ($\mathbf{L}$)
   is the distance transform to the maximum correlation response, representing the per-pixel confidence of target presence.} 
\label{fig:scm-scheme} 
\end{figure}

The maximum of the correlation response is considered as the most likely target position.
The D3S$_2$ output (i.e., segmentation), however, requires specifying a belief of target presence at each pixel. 
Therefore, a target location channel is constructed by computing a (Euclidean) distance transform from the position of the maximum in the correlation map to the remaining pixels in the search region. 
The GEM pathway is shown in Figure~\ref{fig:scm-scheme}.

\subsection{Refinement pathway}  \label{sec:refinement}

\begin{figure}[!t]
\begin{center}
	\includegraphics[width=\linewidth]{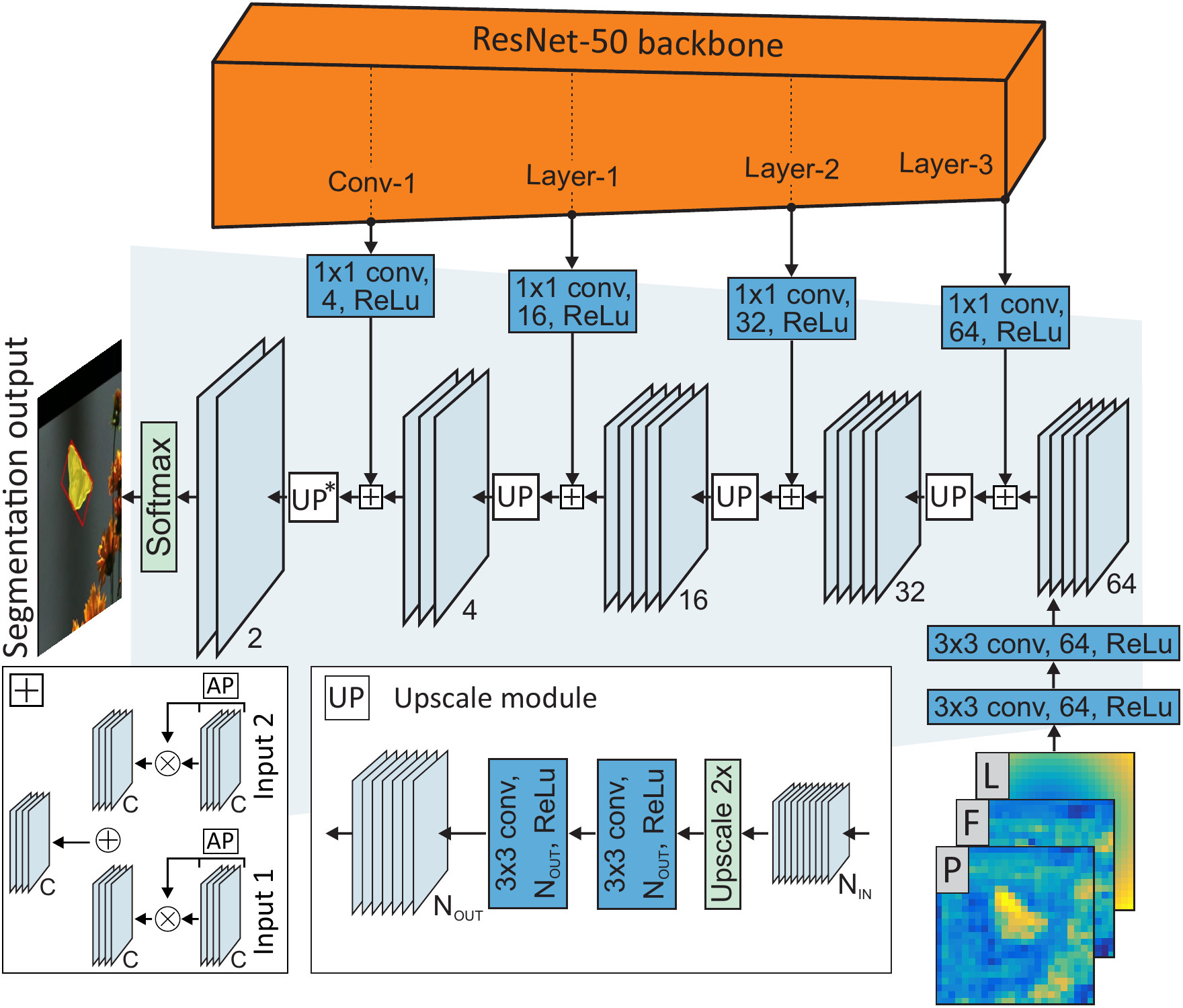}
\end{center}
   \caption{The refinement pathway combines the GIM and GEM channels and gradually upscales  them by using adjusted features from the backbone.
   The AP is an average pooling for channel attention in the feature combination process. The $\oplus$ is summation of two tensors and $\otimes$ represents multiplication of a tensor channels with the elements from a vector.
   The UP$^*$ is a modified UP layer (see~the~text). 
   } 
\label{fig:refinement-architecture} 
\end{figure}

The GIM and GEM pathways provide complementary information about the pixel-level target presence. 
GEM provides a robust, but rather inaccurate estimate of the target region, whereas the output channels from GIM show a greater detail, but are less discriminative (Figure~\ref{fig:intro_figure}). 
Furthermore, the individual outputs are low-resolution due to the backbone encoding. 
A refinement pathway is thus designed to combine the different information channels and upscale the solution into an accurate and detailed segmentation map. 
A network architecture of the refinement pathway is shown in Figure~\ref{fig:refinement-architecture}.

The refinement pathway takes the following inputs: the target location channel ($\mathbf{L}$) from GEM and the foreground similarity and posterior channels ($\mathbf{F}$ and $\mathbf{P}$) from the GIM. 
The channels are concatenated and processed by a $3 \times 3$ convolutional layer followed by a ReLU, resulting in a tensor of 64 channels. 
Three stages of upscaling akin to~\cite{unet_2015,refine_eccv16} are then applied to refine the details by considering the features in different layers computed in the backbone. 
An upscaling stage consists of doubling the resolution of the input channels, 
followed by two $3 \times 3$ convolution layers (each followed by a ReLU).
The resulting channels are summed with the adjusted features from the corresponding backbone layer. 
Specifically, the backbone features are adjusted for the upscaling task by a $1\times 1$ convolution layer, followed by a ReLU.
A channel attention mechanism~\cite{yu_cvpr18} is applied before summing the adjusted backbone feature tensor and the upscaled tensor from the previous refined layer. 
The channel attention operation involves re-weighting the tensor channels by their average pooled values.
The last upscaling stage (which contains only resolution doubling, followed by a single $3\times3$ convolution layer) is followed by a softmax to produce the final segmentation probability map.

\subsection{Scale estimation module}  \label{sec:scale-change-module}

As discussed in the introduction, D3S$_2$ decouples the \textit{inherent target size} (i.e., the scale), responsible for determining the spatial scale for feature extraction in the model update target localization, from the \textit{target visible size} responsible for reporting the target location in the current image. We define the target scale by an axis-aligned bounding box, which also includes all the invisible (potentially occluded) target parts. This bounding box is estimated by the scale estimation module (SEM), described in the following.

\begin{figure}[!t]
\begin{center}
	\includegraphics[width=0.85\linewidth]{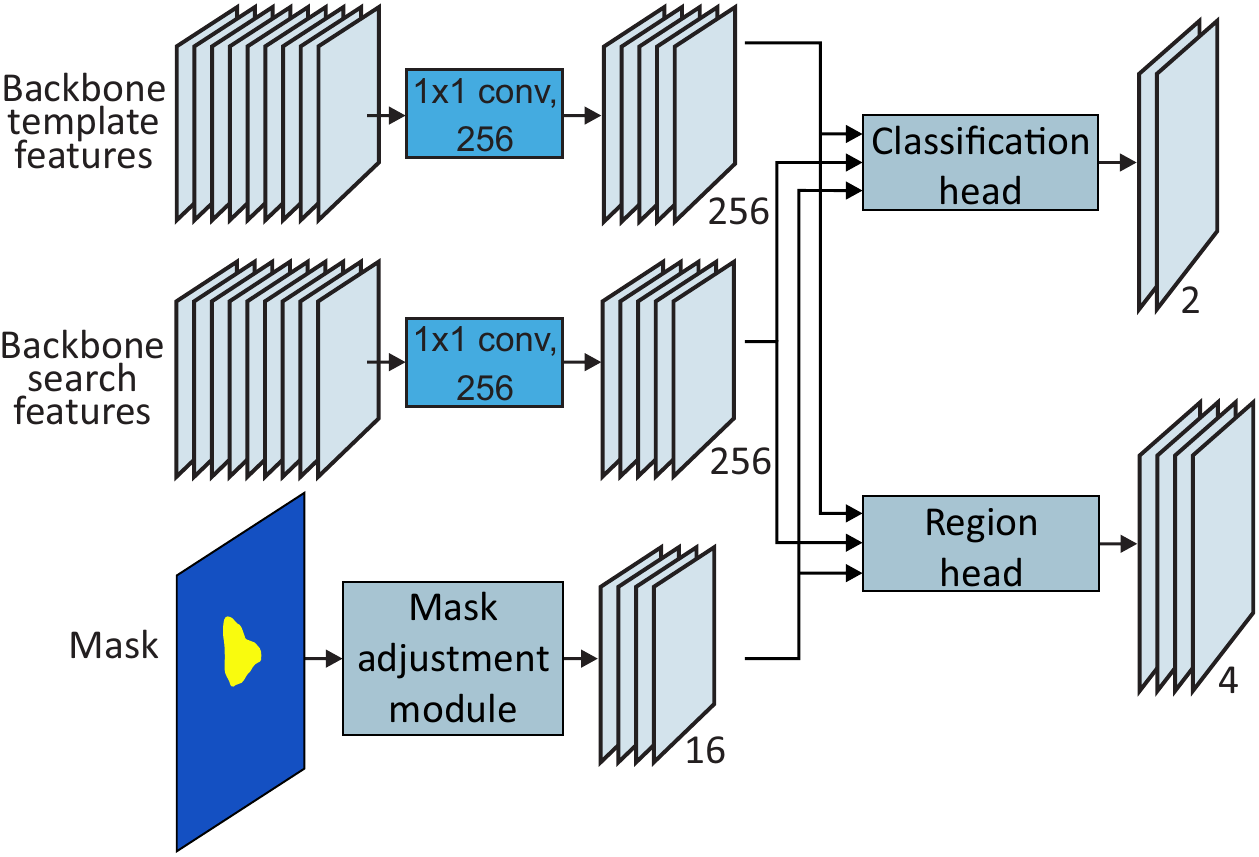}
\end{center}
   \caption{The target scale estimation network (SEM) adjusts the mask by an adjustment module and combines it with the template and search region features to predict per-pixel likelihood of target inherent position along with the size using the target classification head and target region head.}
\label{fig:scale-change-module-scheme} 
\end{figure}

Since the inherent target location may not correspond to the visible target location, SEM predicts several possible target sizes related to different image positions along with corresponding target presence certainty scores (see Figure~\ref{fig:scale-change-module-scheme}).
Note that while similar modules have been proposed in\cite{fcos_iccv2019, siamban_cvpr2020, siamrpn_cvpr2018, siamrpn_cvpr2019}, their usage is different. 
Prior work use such modules for target localization (i.e., tracking output), while D3S$_2$ applies it for robust target scale estimation (not visible size).

\begin{figure}[!t]
\begin{center}
	\includegraphics[width=\linewidth]{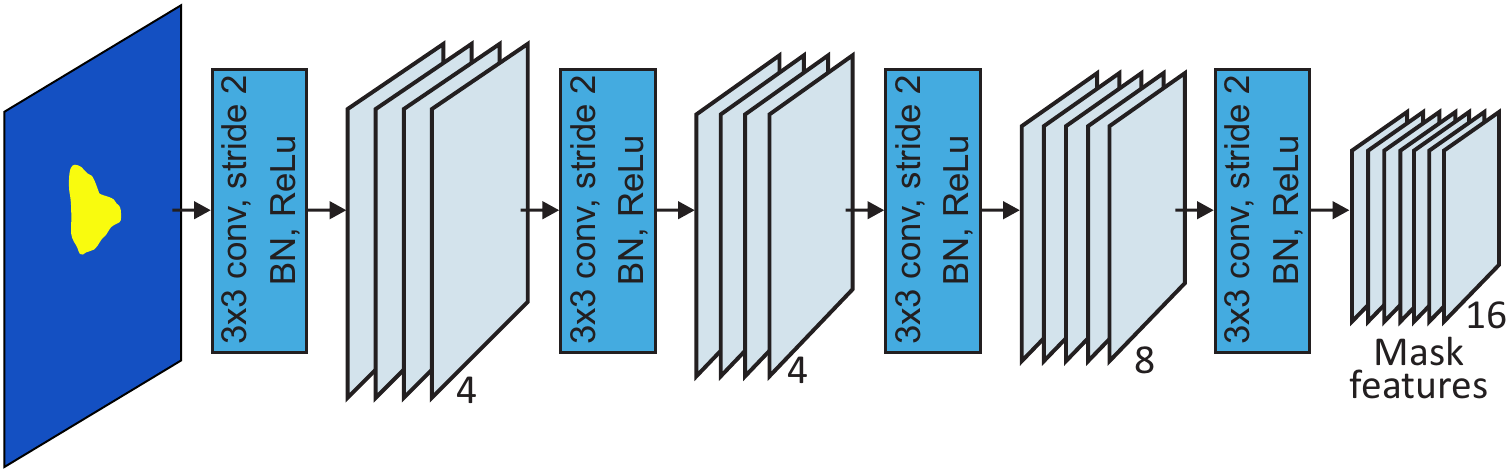}
\end{center}
   \caption{The mask adjustment module extracts mask features adjusted for scale prediction.} 
\label{fig:mask-downsampling} 
\end{figure}

\begin{figure}[!t]
\begin{center}
	\includegraphics[width=\linewidth]{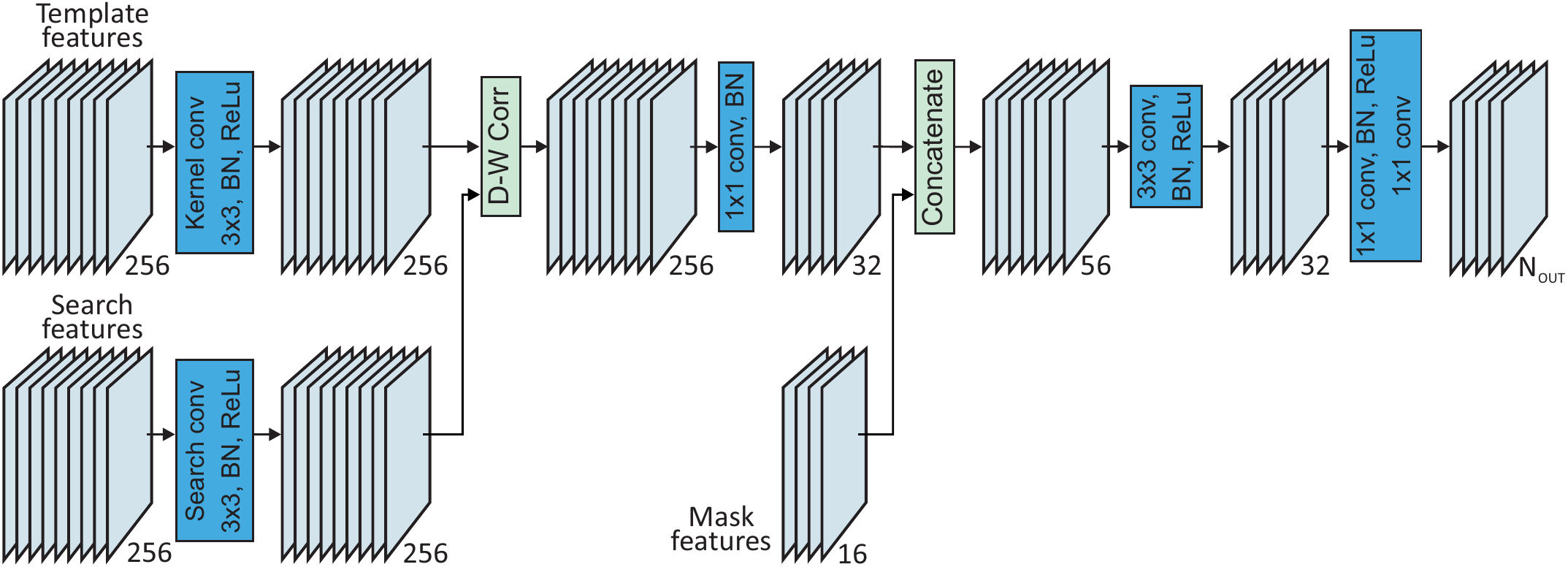}
\end{center}
   \caption{The same architecture is used in the target classification head and the target region head with $N_\mathrm{OUT}=2$ and $N_\mathrm{OUT}=4$ output channels, respectively.
   } 
\label{fig:head-architecture} 
\end{figure}

The input to SEM are the segmentation mask predicted by the refinement pathway, the backbone template features and the search region features. 
The purpose of the mask and the template is to help focusing the network on the visible parts of the target.
The predicted mask is first processed by a mask adjustment module (shown in Figure~\ref{fig:mask-downsampling}), to match the resolution of the backbone features and encode the mask into a form suitable for scale prediction by the downstream network. 
The number of channels in the template and search region features is reduced to 256 by $1\times 1$ convolutions. 
The reduced features and adjusted mask are then combined using the classification and target region heads. 
For simplicity, the two heads have the same architecture (Figure~\ref{fig:head-architecture}), but report a different number of output channels. 

The target classification head predicts two channels, representing per-pixel target and background presence scores. Position of the maximum in the per-pixel softmax between these two channels is the most likely target position, denoted as $\tilde{\mathbf{p}}$.
The target region head predicts four channels $\mathbf{d}_{\mathrm{T}}$, $\mathbf{d}_{\mathrm{B}}$, $\mathbf{d}_{\mathrm{R}}$ and $\mathbf{d}_{\mathrm{L}}$, corresponding to the per-pixel distances to the top, bottom, right and left edges of the target bounding box, respectively. The target width and height ($\tilde{w}, \tilde{h}$), determining the target inherent scale, are thus defined as
\begin{equation} 
    \tilde{w} = \mathbf{d}_{\mathrm{R}}(\tilde{\mathbf{p}}) - \mathbf{d}_{\mathrm{L}}(\tilde{\mathbf{p}}),
\end{equation}
\begin{equation} 
    \tilde{h} = \mathbf{d}_{\mathrm{T}}(\tilde{\mathbf{p}}) - \mathbf{d}_{\mathrm{B}}(\tilde{\mathbf{p}}),
\end{equation}
where $\mathbf{d}_{\mathrm{( \cdot )}}(\tilde{\mathbf{p}})$ is the corresponding channel value at most likely target inherent position $\tilde{\mathbf{p}}$.
  
\section{Discriminative Segmentation Tracker}  \label{sec:tracker}  

This section outlines application of the discriminative segmentation network from Section~\ref{sec:methods} to online general object tracking. 
Given a single supervised training example from the first frame, the network produces target segmentation masks in all the remaining frames. 
However, some applications and most tracking benchmarks require target location represented by a bounding box. For most benchmarks, the bounding box is trivially obtained by fitting an axis-aligned bounding box that tightly fits a segmentation mask. 
The tracking steps are outlined in Section~\ref{sec:diset-tracking}. 

\subsection{Tracking with D3S$_2$}  \label{sec:diset-tracking}
 
\textbf{Initialization.} 
D3S$_2$ is initialized on the first frame using the ground truth target location. 
The GEM and GIM initialization details depend on whether the target ground truth is presented by a bounding box or a segmentation mask. 
If a ground truth bounding box is available, the GEM follows the initialization procedure proposed in~\cite{atom_cvpr19}, which involves training both the dimensionality reduction network and the DCF by backprop on the first frame by considering the region four times the target size. 
On the other hand, if a segmentation mask is available, the ground truth target bounding box is first approximated by an axis-aligned rectangle encompassing the segmented target.

In case a segmentation mask is available, the GIM is initialized by extracting foreground samples from the target mask and background samples from the neighborhood four times the target size. 
However, if only a bounding box is available, an approximate ground truth segmentation mask is constructed first.
Foreground samples are extracted from within the bounding box, while the background samples are extracted from a four times larger neighborhood. A tracking iteration of D3S$_2$ is then run on the initialization region to infer a \textit{proxi} ground truth segmentation mask. 
The final foreground and background samples are extracted from this mask. This process might be iterated a few times (akin to GrabCut~\cite{rother_grabcut2004}), however, we did not observe improvements and chose only a single iteration for initialization speed and simplicity. 
Note that GIM initialization does not involve additional network fine tuning.
 
\textbf{Tracking.} During tracking, when a new frame arrives, a region four times the target size is extracted at previous target location. 
The region is processed by the discriminative segmentation network from Section~\ref{sec:methods} to produce the output segmentation mask. 
The target size is estimated next by the scale estimation module from Section~\ref{sec:scale-change-module}. 
The estimated size is used to update the DCF in GEM by backprop procedure~\cite{atom_cvpr19} and determines the target search region size in the next frame.
In principle the search region size can be defined using (i) the size of the visible part of the target only, 
or (ii) the whole target size, including (potentially occluded) target parts. 
We use the latter option, which means that all training samples are consistently aligned.
An axis-aligned bounding box is fitted to the mask if required by the evaluation protocol.

\section{Experiments}  \label{sec:experiments}

\subsection{Implementation details}  \label{sec:implementation}

The backbone network in D3S$_2$ is composed of the first four layers of ResNet50~\cite{resnet_cvpr16} with octave convolution architecture, 
pre-trained on ImageNet for object classification~\cite{octaveconv_iccv19}. 
The backbone features are extracted from the target search region resized to $384 \times 384$ pixels. 

{\bf Network pre-training.} The GIM pathway and the refinement pathway are pre-trained on 3471 training segmentation sequences from  Youtube-VOS~\cite{yt_vos2018}. 
A training sample is constructed by uniformly sampling a pair of images and the corresponding segmentation masks from the same sequence within a range of 50 frames. 
To increase the diversity of training examples, we include image-based COCO~\cite{coco_eccv14} and randomly perturb an image to construct a training pair. 
Training samples in a batch are sampled from the VOS dataset with probability 80\% and from COCO with probability 20\% to focus on the more challenging training samples from video. 
To increase the robustness to possibly inaccurate GEM localization, the target location channel was constructed by perturbing ground truth locations uniformly from $[-\frac{1}{8}\sigma, \frac{1}{8}\sigma]$, where $\sigma$ is target size.
The network was trained by 64 image pairs batches for 40 epochs with 1000 iterations per epoch using the ADAM optimizer~\cite{adam_2015} with learning rate set to $10^{-3}$ and with 0.2 decay every 15 epochs. 
The training loss was a crossentropy between the predicted and ground truth segmentation mask. The training takes 20 hours on a single GPU.

The  scale estimation module (Section~\ref{sec:scale-change-module}) is trained next on the training subsets of the following datasets:
GoT-10k~\cite{got10k} (8335 sequences), 
Trackingnet~\cite{muller_trackingnet} (30131 sequences)
and LaSoT~\cite{lasot_cvpr19} (1120 sequences), where  
training pairs are sampled with the probability 40\%, 50\% and 10\%, respectively. 
Each batch is composed of 92 pairs of image patches sampled 250 frames apart.

\noindent{\bf Tracking speed.} 
D3S$_2$ is implemented in Pytorch and a single NVidia GTX 1080 GPU was used for network training and evaluation. 
Loading the network to a GPU and initialization of the tracker takes approximately 1.3s, while the average speed is 11.3fps. 
The most time-consuming parts of D3S$_2$ are feature extraction and DCF filter update in GEM. 
The backbone selection and the DCF complexity can thus be considered as hyperparameters that can be adjusted depending on the performance-to-speed trade-offs application requires. 
To demonstrate the impact of these two components, we design a D3S$_2$ with a so-called intermediate configuration, which contains the vanilla ResNet-50 backbone and 64 channels in DCF, denoted as D3S$_{\mathrm{2f}}$. 
This configuration achieves 0.458 EAO on VOT2020, which represents an approximate 11\% performance drop compared to D3S$_2$, but runs at approximately 24 FPS, which is more than two times faster than D3S$_2$. 
The two configurations allow user to choose one of two operating points, one providing a significantly more accurate, but slower, tracker than the original D3S, the other with slightly improved accuracy, but equally fast.

\subsection{Evaluation on Tracking Datasets}  \label{sec:experiments-tracking}

\begin{table*}[!th]
\begin{center}
\begin{tabular}{l r r r r r r r r r r r}
\hline
 & \multicolumn{1}{c}{RPT} & \multicolumn{1}{c}{\textbf{D3S$_2$}} & \multicolumn{1}{c}{Ocean+} & \multicolumn{1}{c}{AlphaRef} & \multicolumn{1}{c}{AFOD} & \multicolumn{1}{c}{LWTL} & \multicolumn{1}{c}{fOcean} & \multicolumn{1}{c}{DET50} & \multicolumn{1}{c}{D3S} & \multicolumn{1}{c}{Ocean} \\
\hline
EAO~$\uparrow$  & \first{0.530} & \second{0.513} & \third{0.491} & 0.482 & 0.472 & 0.463 & 0.461 & 0.441 & 0.439 & 0.430 \\
Acc.~$\uparrow$ & 0.700 & \third{0.714} & 0.685 & \first{0.754} & 0.713  & \second{0.719}  & 0.693  & 0.679 & 0.699 & 0.693 \\
Rob.~$\uparrow$ & \first{0.869} & \second{0.842} & \second{0.842} & 0.777 & 0.795 & 0.798 & 0.803 & 0.787 & 0.769 & 0.754 \\
\hline
\end{tabular}
\end{center}
\caption{Comparison with the top-ten trackers on the VOT2020 Challenge. Note that the D3S denotes the preliminary version~\cite{Lukezic_CVPR_2020} of the proposed tracker.} 
\label{tab:VOT2020-challenge}
\end{table*}

\begin{table*}[!th]
\begin{center}
\begin{tabular}{l r r r r r r r r r r r r r}
\hline
 & \multicolumn{1}{c}{\textbf{D3S$_2$}} & \multicolumn{1}{c}{Ocean} & \multicolumn{1}{c}{SiamMask} & \multicolumn{1}{c}{STM} & \multicolumn{1}{c}{DIMP} & \multicolumn{1}{c}{ATOM} & \multicolumn{1}{c}{Siam++} & \multicolumn{1}{c}{RCNN} & \multicolumn{1}{c}{CSRDCF} & \multicolumn{1}{c}{SiamFC} \\
\hline
EAO~$\uparrow$  & \first{0.513} & \second{0.430} & \third{0.321} & 0.308 & 0.274 & 0.271 & 0.255 & 0.230 & 0.193 & 0.179 \\
Acc.~$\uparrow$ & \second{0.714} & \third{0.639} & 0.624 & \first{0.751} & 0.457 & 0.462 & 0.424 & 0.622 & 0.406 & 0.418 \\
Rob.~$\uparrow$ & \first{0.842} & \second{0.754} & 0.648 & 0.574 & \third{0.740} & 0.734 & 0.730 & 0.484 & 0.582 & 0.502 \\
\hline
\end{tabular}
\end{center}
\caption{VOT2020 -- comparison with the published state-of-the-art trackers.} 
\label{tab:VOT2020-sota}
\end{table*}

\begin{table*}[!th]
\begin{center}
\begin{tabular}{c c c c c c c c c c c c c}
\hline  
 & RCNN & \textbf{D3S$_2$} & DIMP & D3S & Siam++ & ATOM & SiamMask & SiamFCv2 & SiamFC & GOTURN & CCOT & MDNet \\
\hline
AO~$\uparrow$ & \textbf{64.9} & 63.9 & 61.1 & 59.7 & 59.5 & 55.6 & 51.4  & 37.4 & 34.8 & 34.2 & 32.5 & 29.9 \\
SR$_{0.75}$~$\uparrow$ & \textbf{59.7} & 50.9 & 49.2 & 46.2 & 47.9 & 40.2 & 36.6 & 14.4 & 9.8 & 12.4 & 10.7 & \phantom{0}9.9 \\
SR$_{0.5}$~$\uparrow$ & 72.8 & \textbf{73.1} & 71.7 & 67.6 & 69.5 & 63.5 & 58.7 & 40.4 & 35.3 & 37.5 & 32.8 & 30.3 \\
\hline
\end{tabular}
\end{center}
\caption{GOT-10k test set -- comparison with state-of-the-art trackers.} 
\label{tab:got10k}
\end{table*}

\begin{table*}[!th]
\begin{center}
\begin{tabular}{l r r r r r r r r r r r r r}
\hline  
 & RCNN & MAML & Siam++ & \textbf{D3S$_2$} & DIMP & SiamRPN++ & D3S & SiamMask & ATOM & MDNet & CFNet & SiamFC & ECO \\
\hline
AUC~$\uparrow$ & \textbf{81.2} & 75.7 & 75.4 & 75.1 & 74.0 & 73.3 & 72.8 & 72.5 & 70.3 & 60.6 & 57.8 & 57.1 & 55.4 \\
Prec.~$\uparrow$ & \textbf{80.0} & - & 70.5 & 69.9 & 68.7 & 69.4 & 66.4 & 66.4 & 64.8 & 56.5 & 53.3 & 53.3 & 49.2 \\
Prec.$_{\mathrm{N}}$~$\uparrow$ & \textbf{85.4} & 82.2 & 80.0 & 80.4 & 80.1 & 80.0 & 76.8 & 77.8 & 77.1 & 70.5 & 65.4 & 66.3 & 61.8 \\
\hline
\end{tabular}
\end{center}
\caption{TrackingNet test set -- comparison with state-of-the-art trackers.} 
\label{tab:trackingnet}
\end{table*}

\begin{table*}[!th]
\begin{center}
\begin{tabular}{l r r r r r r r r r r r r r r}
\hline  
AUC~$\uparrow$ & RCNN & DIMP & Siam++ & \textbf{D3S$_2$} & MAML & ATOM & SiamRPN++ & D3S & MDNet & VITAL & SiamFC & SSiam & DSiam & ECO \\
\hline
LaSoT & \textbf{65.0} & 56.9 & 54.4 & 54.0 & 52.0 & 51.5 & 49.6 & 49.2 & 39.7 & 39.0 & 33.6 & 33.5 & 33.3 & 32.4 \\
OTB100 & \textbf{70.1} & 68.4 & 68.3 & 67.2 & 70.4 & 66.9 & 69.6 & 60.8 & 67.8 & 68.2 & 58.2 & 62.1 & - & 69.1 \\
OTB100s & 54.9 & 56.6 & - & \textbf{58.0} & - & 56.2 & 52.5 & 52.2 & 49.3 & 49.5 & 41.2 & - & - & 45.6 \\
\hline
\end{tabular}
\end{center}
\caption{Comparison with state-of-the-art trackers on the LaSoT and OTB100 datasets. The OTB100s denotes a set of ten randomly selected OTB100 sequences, which we manually annotated with segmentation masks.} 
\label{tab:lasot-otb}
\end{table*}

D3S$_2$ is evaluated on the most recent tracking dataset VOT2020~\cite{kristan_vot2020}, where targets are annotated by segmentation masks. 
In addition to the VOT2020, the following major bounding-box based short-term tracking datasets are used in the experimental validation: GOT-10k~\cite{got10k}, TrackingNet~\cite{muller_trackingnet}, LaSoT~\cite{lasot_cvpr19} and OTB100~\cite{otb_pami2015}. 
In the following we discuss the results obtained on each of the datasets.

{\bf The VOT2020} dataset consists of 60 sequences. 
Targets are annotated by segmentation masks to enable a more thorough localization accuracy evaluation than on classical datasets with subjective bounding box annotations. 
A new VOT evaluation protocol~\cite{kristan_vot2020} is used. 
The tracker is initialized on the fixed pre-defined initialization frames and let to track forward or backward through the sequence. 
Performance is measured by accuracy (average overlap over successfully  tracked  frames),  robustness (percentage of the successfully tracked frames until tracking failure) and the EAO (expected average overlap), which is a principled combination of the former two measures~\cite{kristan_vot2015}.

D3S$_2$ is compared to the top-ten trackers on the most recent VOT2020 challenge: 
RPT, OceanPlus, AlphaRef, AFOD, LWTL, fastOcean, DET50, Ocean and D3S, which is a preliminary version of the D3S$_2$~\cite{Lukezic_CVPR_2020}.
Results reported in Table~\ref{tab:VOT2020-challenge} show that D3S$_2$ ranks second in terms of the VOT primary measure -- EAO, achieving only a 3\% lower score than top-performing RPT. 
This is a remarkable result since most of the methods are complicated combinations of existing methods from visual tracking, video object segmentation and even object detection.
For example, the top-performing RPT tracker, which substantially outperformed all other entries on VOT2020,
is a combination of several existing methods like RepPoints~\cite{reppoints_iccv19}, DIMP~\cite{dimp_iccv19} and 
a segmentation method from the GIM model from the preliminary D3S~\cite{Lukezic_CVPR_2020} and runs at 4FPS. 
In contrast, D3S$_2$ performs comparably with a much cleaner and lean architecture, running over twice as fast.

In addition to the top-performing trackers on the VOT2020 challenge, we compare the proposed D3S$_2$ to the state-of-the-art published trackers on the VOT2020 dataset. 
The comparison includes the recent trackers Siam R-CNN~\cite{siamrcnn_cvpr2020} (denoted as RCNN), Siam++~\cite{siamfc++_aaai2020}, Ocean~\cite{ocean_eccv2020}, SiamMask~\cite{siammask_cvpr19}, DIMP~\cite{dimp_iccv19} and ATOM~\cite{atom_cvpr19}, 
older tracking baselines CSRDCF~\cite{Lukezic_IJCV_2018} and SiamFC~\cite{siamfc_eccv16} and a state-of-the-art video object segmentation method STM~\cite{stm_iccv19}.
D3S$_2$ is the top-performing method among the published state-of-the-art methods outperforming the most recent second-best performing Ocean by 19\% in EAO. 
Results show that performance increases are due to both, accuracy and robustness, in particular D3S$_2$ outperforms the second-most robust tracker by 12\% and achieves second-most accurate performance, which is 5\% lower than 
the state-of-the-art video-object segmentation method STM~\cite{stm_iccv19}, whose robustness is 32\% lower.

{\bf GOT-10k}~\cite{got10k} is the recent large-scale high-diversity dataset consisting of 10k video sequences with targets annotated by axis-aligned bounding boxes. The trackers are evaluated on 180 test sequences with 84 different object classes and 32 motion patterns, while the rest of the sequences form a training set. 
A tracker is initialized on the first frame and let to track to the end of the sequence. 
Trackers are ranked according to the average overlap, but success rates (SR$_{0.5}$ and SR$_{0.75}$) are reported at two overlap thresholds 0.5 and 0.75, respectively, for detailed analysis\footnote{Success rate denotes percentage of frames where predicted region overlaps with the ground-truth region more than the threshold.}. 
The following top-performing sota trackers are used in comparison~\cite{got10k}: SiamFCv2~\cite{Valmadre_2017_CVPR}, SiamFC~\cite{siamfc_eccv16}, GOTURN~\cite{goturn_eccv2016}, CCOT~\cite{danelljan_eccv2016_ccot}, MDNet~\cite{mdnet_cvpr2016} and 
the most-recent ATOM~\cite{atom_cvpr19}, SiamMask~\cite{siammask_cvpr19}, DIMP~\cite{dimp_iccv19}, Siam R-CNN~\cite{siamrcnn_cvpr2020} (denoted as RCNN) and Siam++~\cite{siamfc++_aaai2020}.
Some of the top-performing sota trackers we use in a comparison utilize the GOT-10k training set for target localization, while the D3S$_2$ is not fine-tuned on the training set, but trains only scale estimation module on the tracking datasets.

Results on GOT-10k are reported in Table~\ref{tab:got10k}.
D3S$_2$ outperforms most of the top-performing state-of-the-art trackers by a large margin in all performance measures, and achieves more than 70\% boost in average overlap compared to the SiamFCv2, which is a top-performer on the~\cite{got10k} benchmark. 
The recent Siam R-CNN outperforms D3S$_2$ by 1.6\% in average overlap, which is mostly due to the strong target re-detection capability. It is also worth pointing out that Siam R-CNN is trained for improved localization on the GoT-10k dataset.
D3S$_2$ outperforms the most recent DIMP and SiamMask trackers by 4.6\% and over 24\% in average overlap, respectively. 
This demonstrates considerable generalization ability over a diverse set of target types without fine-tuning.

{\bf TrackingNet} is another large-scale dataset for training and testing trackers.  
The training set consists of over 30k video sequences, while the testing set contains 511 sequences. 
A tracker is initialized on the first frame and let to track to the end of the sequence. Trackers are ranked according to the area under the success rate curve (AUC), precision (Prec.) and normalized precision (Prec.$_N$). 
The reader is referred to~\cite{muller_trackingnet} for further details about the performance measures.

D3S$_2$ is compared with the top-performing sota trackers according to~\cite{muller_trackingnet}: ECO~\cite{DanelljanCVPR2017}, SiamFC~\cite{siamfc_eccv16}, CFNet~\cite{Valmadre_2017_CVPR}, MDNet~\cite{mdnet_cvpr2016} and 
the most recent sota trackers ATOM~\cite{atom_cvpr19}, SiamMask~\cite{siammask_cvpr19}, SiamRPN++~\cite{siamrpn_cvpr2019}, DIMP~\cite{dimp_iccv19}, Siam R-CNN~\cite{siamrcnn_cvpr2020} (denoted as RCNN), FCOS-MAML~\cite{maml_cvpr2020} (denoted as MAML) and Siam++~\cite{siamfc++_aaai2020}. 
D3S$_2$ significantly outperforms the sota reported in~\cite{muller_trackingnet} as well as most of the recent published state-of-the-art trackers. 
Note that D3S$_2$ is trained for localization by segmentation only on 3471 sequences from YouTube-VOS~\cite{yt_vos2018}, while ATOM, SiamRPN++, DIMP, MAML, Siam R-CNN and Siam++ are fine-tuned on much larger datasets 
(31k, and over 380k sequences, respectively), 
some of them include the TrackingNet training set as well. 
This further supports a considerable generalization capability of D3S$_2$, which is primarily pre-trained for segmentation, not tracking.

{\bf OTB100} is one of the first tracking datasets from 2015, containing 100 tracking sequences with targets annotated by axis-aligned bounding boxes. 
The dataset has largely saturated over the years and more challenging datasets have been developed since. 
Nevertheless, we include evaluation on this dataset since it is still widely used.
A tracker is initialized on the first frame and let to track to the end of the sequence. Trackers are ranked according to the area under the success rate curve (AUC). 

\begin{figure}[!h]
\begin{center}
	\includegraphics[width=\linewidth]{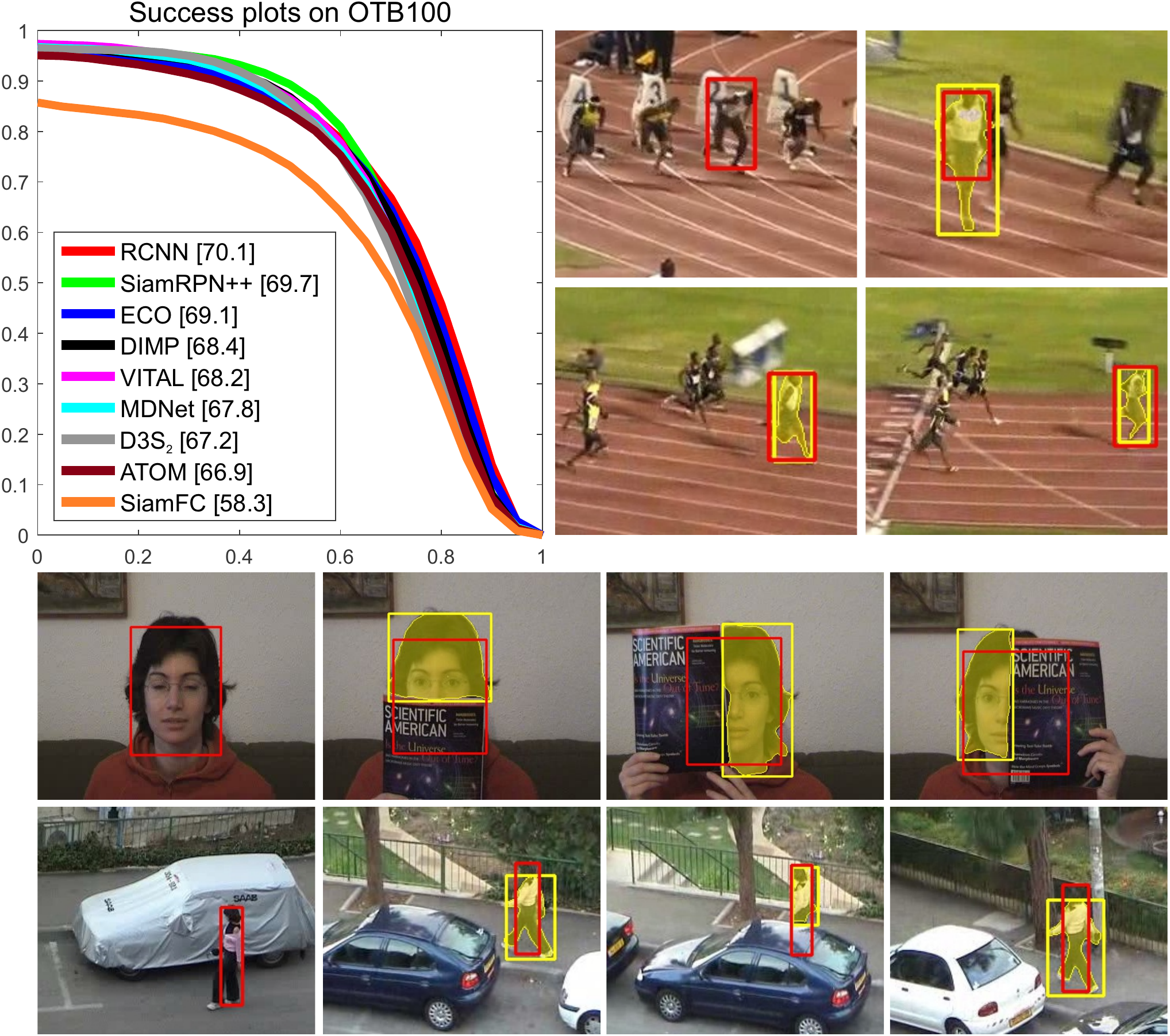}
\end{center}
   \caption{Success plots of the state-of-the-art trackers on OTB100~\cite{otb_pami2015} are shown on the top-left side. Qualitative examples on three sequences from OTB100 are shown top-right and at the bottom. Yellow bounding boxes and segmentation masks denote predictions made by D3S$_2$, while red bounding boxes represent ground-truth. Note the inconsistencies and large ambiguity in ground-truth annotations.} 
\label{fig:otb-examples}
\end{figure}

D3S$_2$ is compared with the trackers which are top performers on OTB: 
ECO~\cite{DanelljanCVPR2017}, SSiam~\cite{structsiam_eccv18}, SiamFC~\cite{siamfc_eccv16}, MDNet~\cite{mdnet_cvpr2016} and 
the recent state-of-the-art trackers VITAL~\cite{vital_cvpr18}, SiamRPN++~\cite{siamrpn_cvpr2019}, ATOM~\cite{atom_cvpr19}, DIMP~\cite{dimp_iccv19}, Siam R-CNN~\cite{siamrcnn_cvpr2020} (denoted as RCNN), Siam++~\cite{siamfc++_aaai2020} and FCOS-MAML~\cite{maml_cvpr2020} (denoted as MAML).
The results are presented in Table~\ref{tab:lasot-otb}. 
D3S$_2$ performs comparably to the top-performing Siam R-CNN with approximately 4\% lower AUC.
Success plots in Figure~\ref{fig:otb-examples} show that all top trackers perform very similarly for overlaps lower than 0.4. 
The main difference comes from the overlaps greater than 0.5. This would imply that some trackers predict the target position more accurately. 
However, we observed that the bounding box annotations on this dataset are ambiguous in certain cases. 
Figure~\ref{fig:otb-examples} shows several examples in which the target location is in fact more accurately estimated by the tracker than the ground truth.
For this reason we randomly selected 10 sequences from OTB100 and manually per-frame re-annotated them with target segmentation masks. 
We re-evaluated the bounding box predictions of the trackers on the segmentation ground truth. 
Tracking performance, measured as the average overlap (AUC), is shown in Table~\ref{tab:lasot-otb}~(OTB100s). 
D3S$_2$ achieves the largest AUC, which shows high accuracy of its predicted bounding boxes.

In addition to comparison of the trackers, we measure the overlap between the OTB bounding box ground-truths and the segmentation annotations. The average overlap on the ten sequences is 0.572, which is comparable to the top-performing trackers.
This result implies that trackers have advanced so much over the last years that the annotation accuracy on this benchmark may no longer differentiate faithfully between the performance of modern trackers.

{\bf LaSoT} is a large scale tracking dataset with total 1400 sequences from 70 diverse categories, where 280 sequences are used for evaluation and each category is equally represented. 
A tracker is initialized at the beginning of the sequence and let to track to the end of the sequence. 
Note that since the targets frequently disappear in the sequences, the dataset is appropriate for evaluation of long-term trackers, i.e., trackers that implement target absence and re-detection capabilities. 
D3S$_2$ is a short-term tracker, but we nevertheless report the performance on this dataset for additional insights.
Tracking performance is measured by the area under the success rate curve (AUC). 
D3S$_2$ is compared to the top-performing trackers according to~\cite{lasot_cvpr19} (MDNet~\cite{mdnet_cvpr2016}, VITAL~\cite{vital_cvpr18}, SiamFC~\cite{siamfc_eccv16}, StructSiam~\cite{structsiam_eccv18}, DSiam~\cite{dsiam_iccv17}, ECO~\cite{DanelljanCVPR2017}) and 
the recent state-of-the-art trackers (DIMP~\cite{dimp_iccv19}, ATOM~\cite{atom_cvpr19}, SiamRPN++~\cite{siamrpn_cvpr2019}, Siam R-CNN~\cite{siamrcnn_cvpr2020} (denoted as RCNN), FCOS-MAML~\cite{maml_cvpr2020} (denoted as MAML) and Siam++~\cite{siamfc++_aaai2020}). 
The results are shown in Table~\ref{tab:lasot-otb}. 

Even though LaSoT is a long-term dataset, the state-of-the-art short-term trackers perform relatively well. The reasons are in large search regions that often cover a substantial portion of the image and in that targets mostly re-appear close to their point of disappearance.
D3S$_2$ ranks fourth, which is a remarkable result for a short-term tracker without target re-detection capability. 
The top-performing Siam R-CNN is a long-term tracker with a strong re-detection capability. 
Some trackers are fine-tuned on LaSoT categories for target localization, while D3S$_2$ uses LaSoT training set for training target size estimation module only. 
This shows great generalization capability of the proposed segmentation-based D3S$_2$.

\subsubsection{Qualitative analysis}

\begin{figure}[!h]
\begin{center}
	\includegraphics[width=\linewidth]{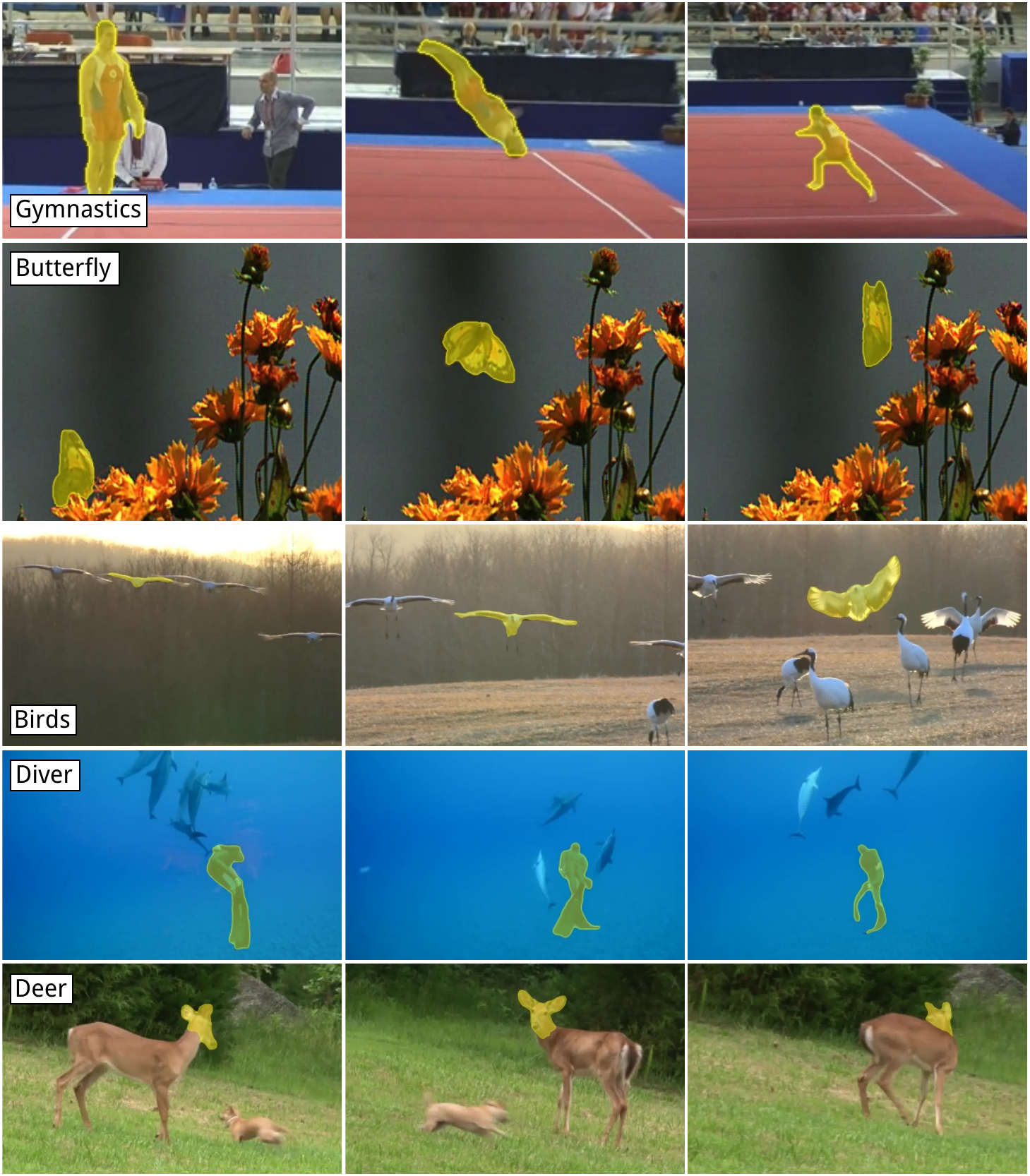}
\end{center}
  \caption{Qualitative examples of D3S$_2$ segmentation masks on video sequences with various challenges: \textit{gymnastics} contains a rotating target with significant scale changes, in  \textit{butterfly} the target shape changes dramatically, \textit{bird} contains several nearby similar deformable targets, \textit{diver} contains a low-contrast underwater scene, and in \textit{deer} only part of the target under partial occlusions is tracked.}
\label{fig:qualitative-D3S}
\end{figure}

The D3S$_2$ performance under challenging conditions is further visualized in  Figure~\ref{fig:qualitative-D3S}. 
The \textit{gymnastics} sequence contains a rotating target that undergoes a scale change in a relatively short time. 
The \textit{butterfly} significantly changes the size and shape, while keeping the inherent scale of features unchanged. 
Owing to decoupling of scale and visible target size, D3S$_2$ faithfully tracks and segments the target in both cases. 
Distractor robustness is demonstrated in the \textit{birds} sequence, which contains several near-by deformable targets. 
D3S$_2$ robustly adapts to the appearance changes of the tracked bird via the GEM module and is not distracted by the other birds, while still delivering an accurate segmentation. 
The \textit{diver} sequence demonstrates robustness to low-contrast scenarios, while the \textit{deer} sequence demonstrates a challenging scenario in which only a part of the target (deer's head) is tracked. 
Even though the head undergoes significant pose changes, and is partially occluded by the deer torso, which is of a similar color, D3S$_2$ adapts to the appearance changes robustly and faithfully segments the non-occluded part of the head. 

\subsection{Ablation Study}  \label{sec:ablation}

An ablation study was performed on VOT2020 using the fixed-reset protocol~\cite{kristan_vot2020} to analyze the contributions of different components of the D3S$_2$ architecture. 
The following variants were considered: 
(i) D3S$_2$ without the geometrically invariant module (D3S$_2^{\overline{GIM}}$); 
(ii) D3S$_2$ without the geometrically constrained Euclidean module (D3S$_2^{\overline{GEM}}$); 
(iii) D3S$_2$ without the channel attention mechanism in refinement pathway (D3S$_2^{\overline{CA}}$);
(iv) D3S$_2$ without the multi-layer perceptron in GIM (D3S$_2^{\overline{MLP}}$);
(v) D3S$_2$ without mask adjustment module (D3S$_2^{\overline{MAM}}$);
(vi) D3S$_2$ without segmentation mask in SEM (D3S$_2^{\overline{M}}$);
(vii) D3S$_2$ without target size estimation module (D3S$_2^{\overline{SEM}}$); 
(viii) D3S$_2$ with 64 channels in DCF, instead of 256 (D3S$_2^{64C}$); 
(ix) D3S$_2$ with a vanilla ResNet50 backbone, instead of ResNet50 with octave convolution and 
(x) preliminary D3S version with the SEM module (D3S$_\mathrm{SEM}$).
We also include the results of the preliminary D3S version~\cite{Lukezic_CVPR_2020} for reference.

\begin{table*}[t]
\begin{center}
\begin{tabular}{l l l l l l l l l l l l l}
 & \multicolumn{1}{c}{D3S$_2$} & \multicolumn{1}{c}{$\mathrm{\overline{M}}$} & \multicolumn{1}{c}{64C} & \multicolumn{1}{c}{$\mathrm{\overline{CA}}$} & \multicolumn{1}{c}{$\mathrm{\overline{MAM}}$} & \multicolumn{1}{c}{$\mathrm{\overline{SEM}}$} & \multicolumn{1}{c}{ResNet50} & \multicolumn{1}{c}{$\mathrm{\overline{MLP}}$} & \multicolumn{1}{c}{$\mathrm{\overline{GEM}}$} & \multicolumn{1}{c}{$\mathrm{\overline{GIM}}$} & \multicolumn{1}{c}{D3S$_\mathrm{SEM}$} & \multicolumn{1}{c}{D3S} \\
\hline
EAO  & 0.513 & 0.508 & 0.504 & 0.497 & 0.496 & 0.492 & 0.479 & 0.468 & 0.463 & 0.439 & 0.475 & 0.439 \\
Acc. & 0.714 & 0.709 & 0.712 & 0.718 & 0.721 & 0.709 & 0.696 & 0.698 & 0.686 & 0.691 & 0.709 & 0.699 \\
Rob. & 0.842 & 0.840 & 0.832 & 0.826 & 0.821 & 0.813 & 0.820 & 0.803 & 0.805 & 0.767 & 0.805 & 0.769 \\
\hline
\end{tabular}
\end{center}
\caption{VOT2020 -- ablation study.
Removing: 
the segmentation mask from the SEM module ({$\mathrm{\overline{M}}$}), 
the channel attention mechanism in refinement pathway ({$\mathrm{\overline{CA}}$}),
the mask adjustment module ({$\mathrm{\overline{MAM}}$}),
the multi-layer perceptron in GIM and replacing it by the averaging top-K values ({$\mathrm{\overline{MLP}}$}) and
the target size estimation module ({$\mathrm{\overline{SEM}}$}).
The {$\mathrm{\overline{GEM}}$} and {$\mathrm{\overline{GIM}}$} denote removing the whole GEM and GIM modules, respectively.
The 64C represents a version with 64 channels in DCF (instead of 256) and ResNet50 uses vanilla ResNet50 backbone, instead of ResNet50 with octave convolution. 
The D3S denotes the preliminary version~\cite{Lukezic_CVPR_2020} of D3S$_2$, while D3S$_\mathrm{SEM}$ denotes the preliminary D3S with SEM module.
} 
\label{tab:ablation}
\end{table*}

Results of the ablation study are presented in Table~\ref{tab:ablation}. 
Removing the channel attention mechanism (D3S$_2^{\overline{CA}}$) from the refinement pathway reduces the tracking performance of the D3S$_2$ by 3\%. 
Inspection of the accuracy/robustness measures, shows that channel attention does not crucially impact the number of failures (robustness), but is responsible for improving the segmentation masks, thus increasing the accuracy.
The improvement comes mostly from the additional channel re-weighting at feature upsampling in the refinement pathway.

To evaluate the quality of mask adjustment in the scale estimation module (SEM), the mask adjustment module is replaced by a bilinear resampling (D3S$_2^{\overline{MAM}}$). The replacement results in a $3\%$ performance drop.
A detailed inspection of the results showed that the main reason for the drop is a reduced robustness due to less accurate target scale estimation in GEM visual model update.

Segmentation mask is removed from the SEM to demonstrate its impact in the scale estimation (D3S$_2^{\overline{M}}$). 
The removal results in an approximate 1\% performance drop, mostly due to the decreased accuracy. 
This results confirms the positive effect of the segmentation mask for robust scale estimation.

To measure the impact of the target scale estimation module, the target scale was estimated
instead by the tracker output (i.e., fitting the bounding box to the segmentation mask). The resulting tracker, D3S$_2^{\overline{SEM}}$ experiences a 4\% performance drop, which is primarily due to increased number of tracking failures (an over 3\% robustness drop).
This strongly supports our hypothesis that robust target scale estimation, which effects mostly the visual model update process, is crucial for robust tracking and speaks in favor of the proposed scale estimation module. 
For a more detailed demonstration of the importance of the SEM module, we perform a per-attribute comparison of D3S$_2^{\overline{SEM}}$ on the LaSoT dataset. 
The results presented in Table~\ref{tab:lasot-per-attribute} show that removing SEM from D3S$_2$ reduces tracking performance under all 14 visual attributes, while the most significant performance drop is observed under fast motion and rotation. This conclusively emphasises the importance of SEM.
To further demonstrate the impact of the SEM, we add it to the preliminary D3S version, and observe increase both in accuracy and robustness and in overall tracking performance improvement by over 8\% in EAO.

The preliminary D3S version~\cite{Lukezic_CVPR_2020} applied a handcrafted average-top-K rule in computing the target foreground/background similarity in GIM. 
To evaluate the contribution of the new multi-layer perceptron method from Section~\ref{sec:sum}, we replace it by the average-top-K rule with $K=3$ as in~\cite{Lukezic_CVPR_2020} (D3S$_2^{\overline{MLP}}$). 
The replacement results in a 9\% performance drop, affecting both accuracy and robustness and confirms strength of the new trainable similarity computation module over the manually-selected top-K rule.

To demonstrate the impact of the larger number of channels in DCF compared to the preliminary version~\cite{Lukezic_CVPR_2020} the version with 64 channels (64C) is included in Table~\ref{tab:ablation}.
Tracking performance is reduced for approximately 2\%, mostly due to the lower robustness. 
This experiment demonstrates that adding more feature channels in GEM reduces tracking speed, but increases the overall tracking performance.

To demonstrate the impact of the backbone, the proposed D3S$_2$ using the ResNet50 with octave convolution is compared to the vanilla ResNet50 backbone. 
Using different backbone results in 6.6\% drop in tracking performance and effects both, accuracy and robustness, which is an expected result since the backbone is shared between GIM and GEM. 
This result confirms that tracking performance of D3S$_2$ may be further improved by using more powerful backbones.

Finally we evaluate the individual contributions of the two visual models GIM and GEM. Removing GEM (D3S$_2^{\mathrm{\overline{GEM}}}$) affects both accuracy and robustness and results in an approximately 10\% overall performance reduction. Note, however, that even with the discriminative model removed, a solid overall tracking performance is still achieved, which indicates that GIM already affords a moderately robust target tracking, while robustness is further increased by GEM. 
Removing GIM (D3S$_2^{\mathrm{\overline{GIM}}}$), however, reduces the accuracy and robustness by 3\% and 9\%, respectively, causing a 14\% overall performance drop. This means that GIM is crucial for segmentation accuracy and localization robustness.

\begin{figure}[!t]
\begin{center}
	\includegraphics[width=\linewidth]{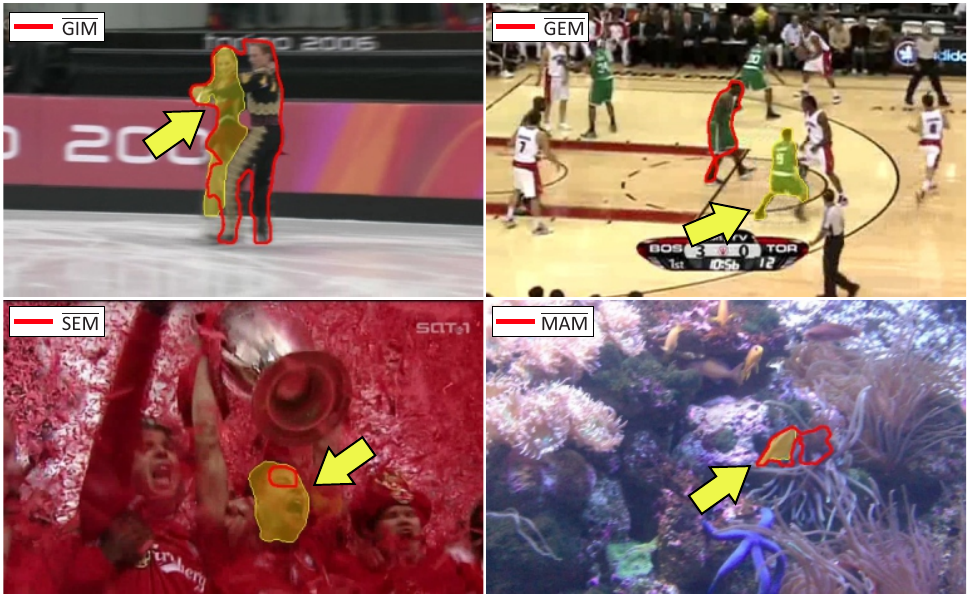}
\end{center}
  \caption{Tracking results of D3S$_2$ (yellow mask) compared to  D3S$_2$ variations (red) with the individual modules, GIM, GEM, SEM and MAM, removed. The arrow indicates the tracked target.} 
\label{fig:ablation-qualitative}
\end{figure}

To gain further insights into importance of each module, we report qualitative comparison of four D3S$_2$ variants, each variant with one of the modules excluded, in Figure~\ref{fig:ablation-qualitative}. 
Removing the geometrically invariant model ($\mathrm{\overline{GIM}}$) leads to over-segmentation in situations with nearby similar objects due to missing local target-specific and background-specific representation. 
Removing the geometrically constrained model ($\mathrm{\overline{GEM}}$) reduces capability of learning a target-specific discriminative model, which leads to drifting to a wrong target. 
Removing the scale estimation module ($\mathrm{\overline{SEM}}$) leads to incorrectly estimating the target size; in Figure~\ref{fig:ablation-qualitative}, for example, the tracker region collapses from a person’s face to their eye. 
Removing the mask adjustment module ($\mathrm{\overline{MAM}}$) reduced the accuracy of feature scale estimation, which leads to an over-segmentation in the case of fish tracking in Figure~\ref{fig:ablation-qualitative}.

\begin{table*}[t]
\begin{center}
\begin{tabular}{r l l l l l l l l l l l l l l l}
 & \multicolumn{1}{c}{AR} & \multicolumn{1}{c}{LR} & \multicolumn{1}{c}{OV} & \multicolumn{1}{c}{FM} & \multicolumn{1}{c}{FO} & \multicolumn{1}{c}{SV} & \multicolumn{1}{c}{VC} & \multicolumn{1}{c}{BC} & \multicolumn{1}{c}{RO} & \multicolumn{1}{c}{CM} & \multicolumn{1}{c}{MB} & \multicolumn{1}{c}{DE} & \multicolumn{1}{c}{PO} & \multicolumn{1}{c}{IL} & \multicolumn{1}{c}{Overall} \\
\hline
D3S$_2$ & 51.9 & 44.9 & 47.8 & 41.4 & 41.3 & 53.8 & 50.7 & 47.4 & 52.0 & 57.6 & 50.9 & 56.3 & 49.2 & 56.5 & 54.0 \\
$\mathrm{\overline{SEM}}$ & 47.1 & 41.6 & 43.4 & 36.2 & 38.0 & 48.9 & 45.0 & 43.1 & 45.4 & 53.2 & 47.8 & 50.5 & 44.0 & 50.7 & 49.2 \\
\hline
\end{tabular}
\end{center}
\caption{
Comparison of the D3S$_2$ and its variant without the SEM module ($\mathrm{\overline{SEM}}$) on the LaSoT dataset wrt. the 14 visual attributes: aspect ratio change (AR), low resolution (LR), Out-of-view (OV), fast motion (FM), full occlusion (FO), scale variation (SV), viewpoint change (VC), background clutter (BC), rotation (RO), camera motion (CM), motion blur (MB), deformation (DE), partial occlusion (PO) and illumination variation (IL).
} 
\label{tab:lasot-per-attribute}
\end{table*}

\subsection{Evaluation on Segmentation Datasets}  \label{sec:experiments-segmentation}

Segmentation capabilities of D3S$_2$ were analyzed on two popular video object segmentation benchmarks DAVIS16~\cite{davis16} and DAVIS17~\cite{davis17}.
Under the DAVIS protocol, the segmentation algorithm is initialized on the first frame by a segmentation mask. The algorithm is then required to output the segmentation mask for all the remaining frames in the video. Performance is evaluated by two measures averaged over the sequences: mean Jaccard index ($\mathcal{J_{M}}$) and mean F-measure ($\mathcal{F_{M}}$).
Jaccard index represents a per-pixel intersection over union between the ground-truth and the predicted segmentation mask. The F-measure is a harmonic mean of precision and recall calculated between the contours extracted from the ground-truth and the predicted segmentation masks. 
For further details on these performance measures, the reader is referred to~\cite{davis16,martin_pami2004}.

D3S$_2$ is compared to several sota video object segmentation methods specialized to the DAVIS challenge setup: OSVOS~\cite{osvos_cvpr2017}, OnAVOS~\cite{onavos_bmvc2017}, OSMN~\cite{osmn_cvpr2018}, FAVOS~\cite{favos_cvpr2018}, VM~\cite{videomatch_eccv2018}, PML~\cite{blazingly_fast_cvpr18} and 
A-GAME~\cite{Johnander_CVPR2019}. 
In addition, we include the most recent segmentation-based tracker SiamMask~\cite{siammask_cvpr19}, which is the only published method that performs well on both, short-term tracking as well as on video object segmentation benchmarks. 
We include also the preliminary D3S version for a complete comparison.

Results are shown in Table~\ref{tab:davis}. D3S$_2$ performs on par with most of the video  object segmentation top performers on DAVIS. 
Compared to top performer on DAVIS2016, the performance of D3S$_2$ is 
14\% and 17\% lower in the average Jaccard index and the F-measure, respectively.
This is quite remarkable, considering that D3S$_2$ is more than 100 times faster.
On DAVIS2017 
the performance drop is 18\% in Jaccard index and 14\% in F-measure compared to the top-performer A-GAME.
In addition, A-GAME was trained on YT-VOS and DAVIS training set, which
demonstrates a considerable generalization capability of D3S$_2$, which was not trained on the DAVIS training set.
Furthermore, D3S$_2$ delivers a comparable segmentation accuracy as pure segmentation methods ASMN and PML, while being orders of magnitude faster, which is particularly important for many video editing applications.
 
\begin{figure}[!t]
\begin{center}
	\includegraphics[width=\linewidth]{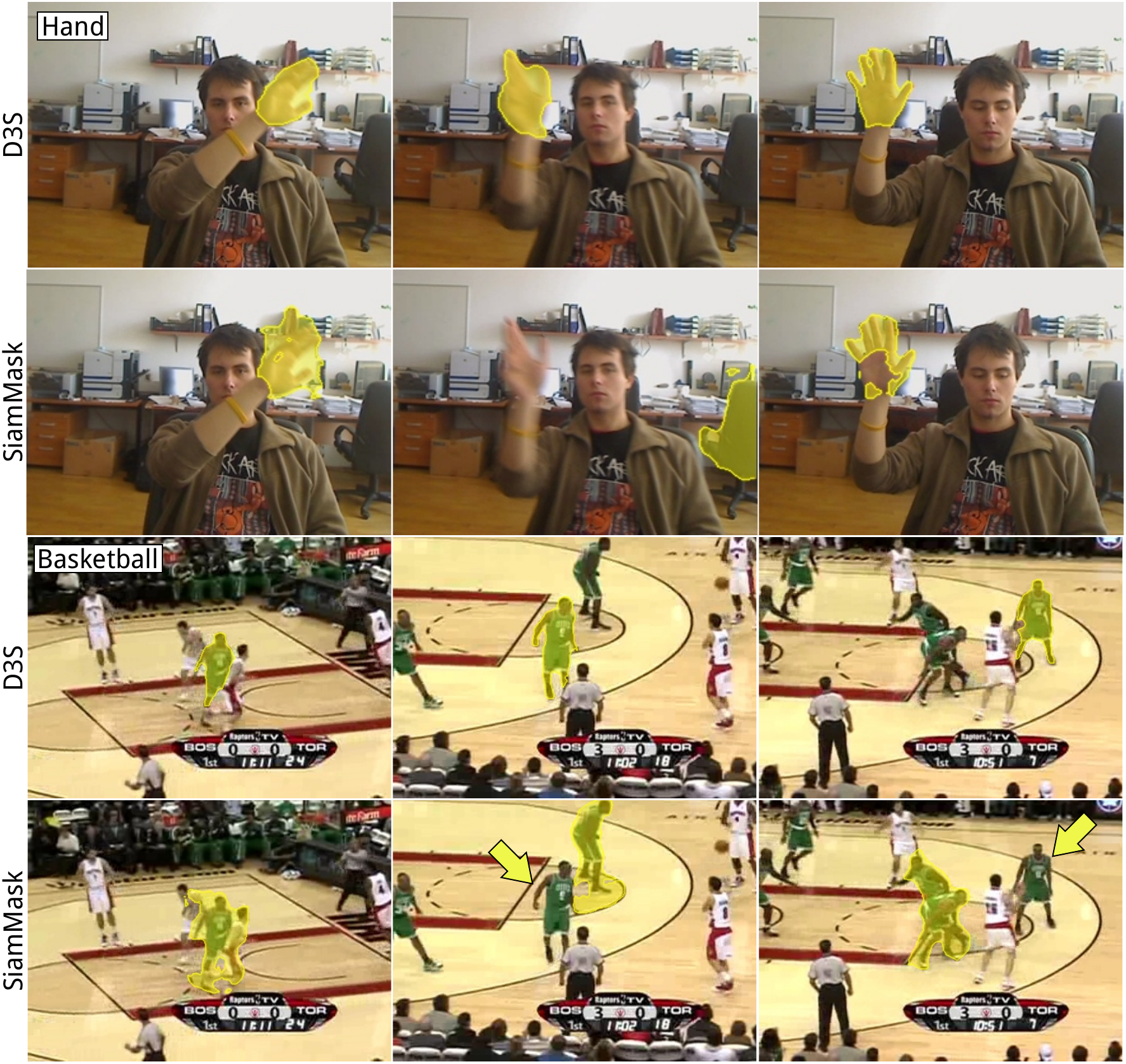}
\end{center}
  \caption{D3S$_2$ vs. SiamMask segmentation quality. 
  \textit{Hand}: the rigid template in SiamMask fails on a deforming target, the GIM model in D3S$_2$ successfully tracks despite a significant deformation. 
  \textit{Basketball}: SiamMask drifts to a similar object, while D3S$_2$ leverages the discriminative learning in GEM to robustly track the selected target. The yellow arrow denotes the target which should be tracked.} 
\label{fig:qualitative-segmentation-comparison}
\end{figure}

D3S$_2$ also outperforms the only tracking-by-segmentation method SiamMask with respect to all measures. On average the segmentation is improved by over 3\% in the Jaccard index and the contour accuracy-based F-measure. See Figure~\ref{fig:qualitative-segmentation-comparison} for further qualitative comparison of D3S$_2$ and SiamMask on challenging targets. 

The performance of D3S$_2$ is slightly lower on video object segmentation compared to its preliminary version D3S~\cite{Lukezic_CVPR_2020}. Videos in video object segmentation benchmarks are fairly short (on average less than 100 frames) with targets quite large, which means that already fairly robust trackers will not fail and performance measures will be dominated by the accuracy of object segmentation. The segmentation from D3S appears slightly better for the large objects. However, in more challenging segmentation scenarios with small objects undergoing significant appearance changes on long sequences (see Table~\ref{tab:VOT2020-challenge}), D3S$_2$ significantly outperforms D3S both in robustness and segmentation accuracy.
Thus, compared to its preliminary version, D3S$_2$ only slightly sacrifices the segmentation accuracy of large objects on relatively short video object segmentation sequences for significant robustness and segmentation performance boosts in challenging tracking scenarios.

\begin{table}[t]
\begin{center}
\begin{tabular}{l c c c c r }
\hline  
 & $\mathcal{J_{M}}^{16}$ & $\mathcal{F_{M}}^{16}$ & $\mathcal{J_{M}}^{17}$ & $\mathcal{F_{M}}^{17}$ & FPS \\ 
\hline
D3S$_2$                          & 74.2 & 70.6 & 55.9 & 62.2 & 11.0  \\
D3S~\cite{Lukezic_CVPR_2020}     & 75.4 & 72.6 & 57.8 & 63.8 & 25.0  \\
SiamMask~\cite{siammask_cvpr19}  & 71.7 & 67.8 & 54.3 & 58.5 & 55.0  \\
OnAVOS~\cite{onavos_bmvc2017}    & 86.1 & 84.9 & 61.6 & 69.1 & 0.1 \\
FAVOS~\cite{favos_cvpr2018}      & 82.4 & 79.5 & 54.6 & 61.8 & 0.8 \\
A-GAME~\cite{Johnander_CVPR2019} & 82.0 & 82.2 & 67.2 & 72.2 & 14.0 \\
VM~\cite{videomatch_eccv2018}    & 81.0 &   -  & 56.6 &   -  & 3.1 \\
OSVOS~\cite{osvos_cvpr2017}      & 79.8 & 80.6 & 56.6 & 63.9 & 0.1 \\
PML~\cite{blazingly_fast_cvpr18} & 75.5 & 79.3 &   -  &   -  & 3.6 \\
OSMN~\cite{osmn_cvpr2018}        & 74.0 & 72.9 & 52.5 & 57.1 & 8.0   \\
\hline
\end{tabular}
\end{center}
\caption{State-of-the-art comparison on the DAVIS16 and DAVIS17 segmentation datasets. Average Jaccard index and F-measure are denoted as $\mathcal{J_{M}}^{16}$ and $\mathcal{F_{M}}^{16}$ on DAVIS16 dataset and $\mathcal{J_{M}}^{17}$ and $\mathcal{F_{M}}^{17}$ on DAVIS17 dataset, respectively.} 
\label{tab:davis}
\end{table}

\section{Conclusion}  \label{sec:conclusion}

A deep single-shot discriminative segmentation tracker -- D3S$_2$ -- was introduced. The tracker leverages two  models from the extremes of the spectrum: a geometrically invariant model and a geometrically restricted Euclidean model. 
The two models localize the target in parallel pathways and complement each other to achieve high segmentation accuracy of deformable targets and robust discrimination of the target from distractors. 
D3S$_2$ decouples the target scale estimation, which is related to the inherent target size, from the target region, which is related to the reported target position. The scale is robustly estimated by a new module and is used in model update and in setting the search region scale. 
The target position is reported as a segmentation mask or by a fitted axis-aligned bounding box if required by the evaluation method.
The end-to-end trainable network architecture is the first single-shot pipeline with online adaptation that tightly connects discriminative tracking with accurate segmentation.

D3S$_2$ outperforms most state-of-the-art trackers on the VOT2020, GOT-10k and TrackingNet benchmarks and performs on par with top trackers on OTB100 and LaSoT, despite the fact that some of the tested trackers were re-trained for specific datasets. 
In contrast, D3S$_2$ was trained once on Youtube-VOS for segmentation only, the scale estimation module is trained on the public tracking training sequences, and the same version was used in \textit{all benchmarks}.
To cover a large range of potential applications, we present two versions of the tracker with different performance-to-speed trade-offs. 
The one provides a significantly more accurate, but slower performance, denoted as D3S$_2$ and the other named D3S$_{\mathrm{2f}}$ with 11\% lower tracking performance, while being more than two times faster. 
We envision future extensions of the presented architecture. One such extension might be introduction of the recently proposed covariant backbone features from~\cite{Gupta_2021_CVPR} to further improve invariance of the geometrically constrained Euclidean module under target rotation.

Tests on DAVIS16 and DAVIS17 segmentation benchmarks show performance close to top segmentation methods while running much faster than most.
D3S$_2$ significantly outperforms recent top segmentation tracker SiamMask on all bechmarks in all performance measures and contributes towards narrowing the gap between two, currently separate, domains of short-term tracking and video object segmentation, thus blurring the boundary between the two.

\ifCLASSOPTIONcompsoc
  \section*{Acknowledgments}
\else
  \section*{Acknowledgment}
\fi

This work was supported by Slovenian research agency program 
P2-0214 
and projects 
J2-2506 
and 
J2-9433. 
A. Lukežič was financed by the Young researcher program of the ARRS.
J. Matas was supported by the Czech Science Foundation grant GA18-05360S. 
We would also like to thank Žiga Sajovic and Tilen Nedanovski for their technical contributions and Jer Pelhan for his contribution on the OTB sequences segmentation.

\vskip .5cm

\noindent \copyright 20XX IEEE. Personal use of this material is permitted. 
Permission from IEEE must be obtained for all other uses, in any current or future media, 
including reprinting/republishing this material for advertising or promotional purposes, 
creating new collective works, for resale or redistribution to servers or lists, 
or reuse of any copyrighted component of this work in other works.
The original version of the paper is available at: \\ 
\url{https://ieeexplore.ieee.org/document/9662200}

\ifCLASSOPTIONcaptionsoff
  \newpage
\fi

\bibliographystyle{IEEEtran}
\bibliography{bib}

\begin{IEEEbiography}[{\includegraphics[width=1in,height=1.25in,clip,keepaspectratio]{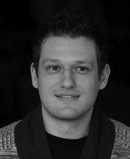}}]
{Alan Lukežič} 
received the Ph.D. degree from the Faculty of Computer and Information Science, University of Ljubljana, Slovenia in 2021. 
He is currently with the Visual Cognitive Systems Laboratory, Faculty of Computer and Information Science, University of Ljubljana, as a Teaching Assistant and a Researcher. 
His research interests include computer vision, data mining and machine learning.
\end{IEEEbiography}

\begin{IEEEbiography}[{\includegraphics[width=1in,height=1.25in,clip,keepaspectratio]{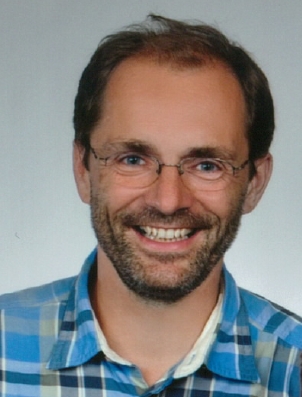}}]
{Jiří Matas} is a professor at the Center for Machine Perception,
Czech Technical University in Prague. He holds a PhD degree from the University
of Surrey, UK (1995). He has published more than 200 papers in refereed journals
and conferences. His publications have approximately 34000 citations in Google Scholar and 13000 in the Web of Science. His h-index is 65 (Google scholar) and 43 (Clarivate Analytics Web of Science)   
respectively. He received the best paper prize at the British Machine Vision
Conferences in 2002 and 2005, at the Asian Conference on Computer Vision
in 2007 and at Int. Conf. on Document analysis and Recognition in 2015.
 He is on the editorial board of IJCV and was an Associate Editor-in-Chief of IEEE T. PAMI. 
His research interests include visual tracking, object recognition,
image matching and retrieval, sequential pattern recognition, and RANSAC-
type optimization methods.
\end{IEEEbiography}

\begin{IEEEbiography}[{\includegraphics[width=1in,height=1.25in,clip,keepaspectratio]{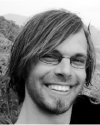}}]
{Matej Kristan} 
received a Ph.D from the Faculty of Electrical Engineering, University of Ljubljana in 2008. 
He is a full professor and a vice chair of the department of artificial intelligence at the Faculty of Computer and Information Science. He is president of the IAPR Slovenian pattern recognition society and an Associate Editor of IJCV.
His research interests include visual object tracking, anomaly detection, object detection and segmentation, perception methods for autonomous boats and physics-informed machine-learning.
\end{IEEEbiography}

\end{document}